%% file: main.tex
\documentclass{article}
\usepackage[numbers]{natbib}
\pdfoutput=1
\usepackage[preprint]{neurips_2024}

\usepackage{times}
\usepackage{latexsym}

\usepackage[T1]{fontenc}

\usepackage[utf8]{inputenc}

\usepackage{microtype}

\usepackage{inconsolata}

\usepackage{multirow}
\usepackage[normalem]{ulem}

\usepackage{xcolor}
\definecolor{PlanningColor}{RGB}{77, 138, 206} % a shade of blue for Planning
\definecolor{ActionColor}{RGB}{175, 76, 34}    % a shade of green for Action
\definecolor{CriticColor}{RGB}{95, 158, 127}   % a shade of red for Critic
\definecolor{ErrorColor}{RGB}{213, 44, 31}     % a shade of red for Critic

\usepackage[linesnumbered,ruled]{algorithm2e}
\usepackage{amsmath}
\usepackage{url}
\usepackage{hyperref}
\definecolor{citecolor}{HTML}{0071bc}
\hypersetup{
    linkcolor=red, % Color of internal links (like \ref)
    colorlinks=true, % Ensures the link is actually colored
    citecolor=citecolor,
}

\usepackage{cleveref}
\usepackage{booktabs}
\usepackage{varwidth}
\usepackage{paralist}
\usepackage{enumerate}
\usepackage{amssymb}
\usepackage{pifont}
\usepackage{bm}
\usepackage{amsmath}
\usepackage{multirow}
\usepackage{caption}
\usepackage[pdftex]{graphicx}
\usepackage{subfigure}

\usepackage{enumitem}
\usepackage{listings}
\usepackage{mdframed}
\usepackage{fvextra}     % For enhanced verbatim environments (with page break support)
\usepackage{bera}         % Optional: For a nice mono-spaced font

\definecolor{mytextcolor}{rgb}{0.25,0.5,0.5}  % Teal for general text
\definecolor{mykeywordcolor}{rgb}{0.1, 0.1, 0.8} % Blue for keywords
\definecolor{mybracketcolor}{RGB}{0,191,255} % Navy blue for brackets
\definecolor{mytagcolor}{rgb}{0.8, 0.1, 0.1}     % Red for HTML tags
\definecolor{myattrcolor}{rgb}{0.1, 0.5, 0.1}    % Green for HTML attributes
\definecolor{myvaluecolor}{rgb}{0.6, 0.2, 0.8}   % Purple for attribute values
\definecolor{mybulletcolor}{RGB}{205, 92, 92}  % Gray for bullet points and numbers

\newmdenv[
  backgroundcolor=gray!10,  % Light gray background for the frame
  linecolor=gray!80,        % Gray border
  linewidth=1pt,            % Border thickness
  roundcorner=5pt,          % Rounded corners
  innertopmargin=5pt,       % Inner margin at the top
  innerbottommargin=5pt,    % Inner margin at the bottom
  innerleftmargin=5pt,      % Inner margin on the left
  innerrightmargin=5pt,     % Inner margin on the right
  skipabove=10pt,           % Space above the box
  skipbelow=10pt,           % Space below the box
  splitbottomskip=5pt,      % Space when splitting across pages
  splittopskip=5pt,         % Space when splitting across pages
  nobreak=false,            % Allow page breaks
  frametitle={Text Block}, % Title of the frame
]{textblock}

\DefineVerbatimEnvironment{Verbatim}{Verbatim}{breaklines, breaksymbolleft={}}

\definecolor{jsonkey}{rgb}{0.0, 0.5, 0.0}     % Green for keys
\definecolor{jsonstring}{rgb}{0.6, 0.0, 0.0}  % Red for string values
\definecolor{jsonnumber}{rgb}{0.0, 0.0, 0.6}  % Blue for numbers
\definecolor{background}{rgb}{0.95, 0.95, 0.95} % Light gray background

\definecolor{eclipseStrings}{RGB}{0,191,255}       % Blue for comments/keys
\definecolor{eclipseKeywords}{RGB}{205, 92, 92}      % Dark red for strings
\colorlet{numb}{magenta!60!black}                 % Magenta for numbers

\lstdefinelanguage{json}{
    basicstyle=\normalfont\ttfamily\footnotesize\color{mytextcolor},              % Monospaced font
    commentstyle=\color{eclipseStrings},          % Color for keys (comments in this setup)
    stringstyle=\color{eclipseKeywords},          % Color for string values
    numbers=left,                                 % Line numbers on the left
    numberstyle=\scriptsize,                      % Small line numbers
    stepnumber=1,                                 % Show every line number
    numbersep=5pt,                                % Space between numbers and code
    showstringspaces=false,                       % Do not display spaces in strings
    breaklines=true,                              % Automatically break long lines
    frame=lines,                                  % Adds lines around the block
    backgroundcolor=\color{gray!10},              % Light gray background
    string=[s]{"}{"},                             % String delimiters are double quotes
    comment=[l]{:\ "},                            % Treat key-value pairs as comments
    morecomment=[l]{:"},                          % Additional comment rule for key-value pairs
    literate=
        *{0}{{{\color{numb}0}}}{1}                % Highlight numbers
         {1}{{{\color{numb}1}}}{1}
         {2}{{{\color{numb}2}}}{1}
         {3}{{{\color{numb}3}}}{1}
         {4}{{{\color{numb}4}}}{1}
         {5}{{{\color{numb}5}}}{1}
         {6}{{{\color{numb}6}}}{1}
         {7}{{{\color{numb}7}}}{1}
         {8}{{{\color{numb}8}}}{1}
         {9}{{{\color{numb}9}}}{1}
}

\newmdenv[
  backgroundcolor=gray!10,  % Light gray background for the frame
  linecolor=gray!80,        % Gray border
  linewidth=1pt,            % Border thickness
  roundcorner=5pt,          % Rounded corners
  innertopmargin=5pt,       % Inner margin at the top
  innerbottommargin=5pt,    % Inner margin at the bottom
  innerleftmargin=5pt,      % Inner margin on the left
  innerrightmargin=5pt,     % Inner margin on the right
  skipabove=10pt,           % Space above the box
  skipbelow=10pt,           % Space below the box
  splitbottomskip=5pt,      % Space when splitting across pages
  splittopskip=5pt,         % Space when splitting across pages
  nobreak=false,            % Allow page breaks
  frametitle={JSON Example}, % Title of the frame
]{jsonframe}

\crefname{equation}{Eq.}{Eq.}
\crefname{section}{Section}{Sections}
\crefname{subsection}{Section}{Sections}
\crefname{subsubsection}{Section}{Sections}
\crefname{figure}{Figure}{Figures}
\crefname{table}{Table}{Tables}
\crefname{subfigure}{Figure}{Figures}
\crefname{algocf}{Algorithm}{Algorithms}

\makeatletter
\DeclareRobustCommand\onedot{\futurelet\@let@token\@onedot}
\def\@onedot{\ifx\@let@token.\else.\null\fi\xspace}

\title{The Dawn of GUI Agent: A Preliminary \\ Case Study with Claude 3.5 Computer Use}

\author{
  Siyuan Hu, Mingyu Ouyang, Difei Gao, Mike Zheng Shou \\
  Show Lab, National University of Singapore
}

\begin{document}

\maketitle

% Abstract
\input{sections/0-abstract}

% Main sections
\input{sections/1-intro}
\input{sections/2-related_work}
\input{sections/3-method}
\input{sections/4-CUAE}

\input{sections/5-discussion}
\input{sections/6-conclusion}

% References
\bibliography{main}
\bibliographystyle{unsrt}

\end{document}

%% file: sections/0-abstract.tex
\begin{abstract}

The recently released model, \textbf{Claude 3.5 Computer Use}, stands out as the first frontier AI model to offer computer use in public beta as a graphical user interface (GUI) agent. As an early beta, its capability in the real-world complex environment remains unknown. In this case study to explore Claude 3.5 Computer Use, we curate and organize a collection of carefully designed tasks spanning a variety of domains and software. Observations from these cases demonstrate Claude 3.5 Computer Use’s unprecedented ability in end-to-end language to desktop actions. Along with this study, we provide an out-of-the-box agent framework for deploying API-based GUI automation models with easy implementation. Our case studies aim to showcase a groundwork of capabilities and limitations of Claude 3.5 Computer Use with detailed analyses and bring to the fore questions about planning, action, and critic which must be considered for future improvement. We hope this preliminary exploration will inspire future research into the GUI agent community. \textbf{All the test cases} in the paper can be tried through the project: \url{https://github.com/showlab/computer_use_ootb}.

\end{abstract}

%% file: sections/1-intro.tex
\begin{figure*}[!h]
\centering
\centerline{\includegraphics[width=1.1\linewidth]{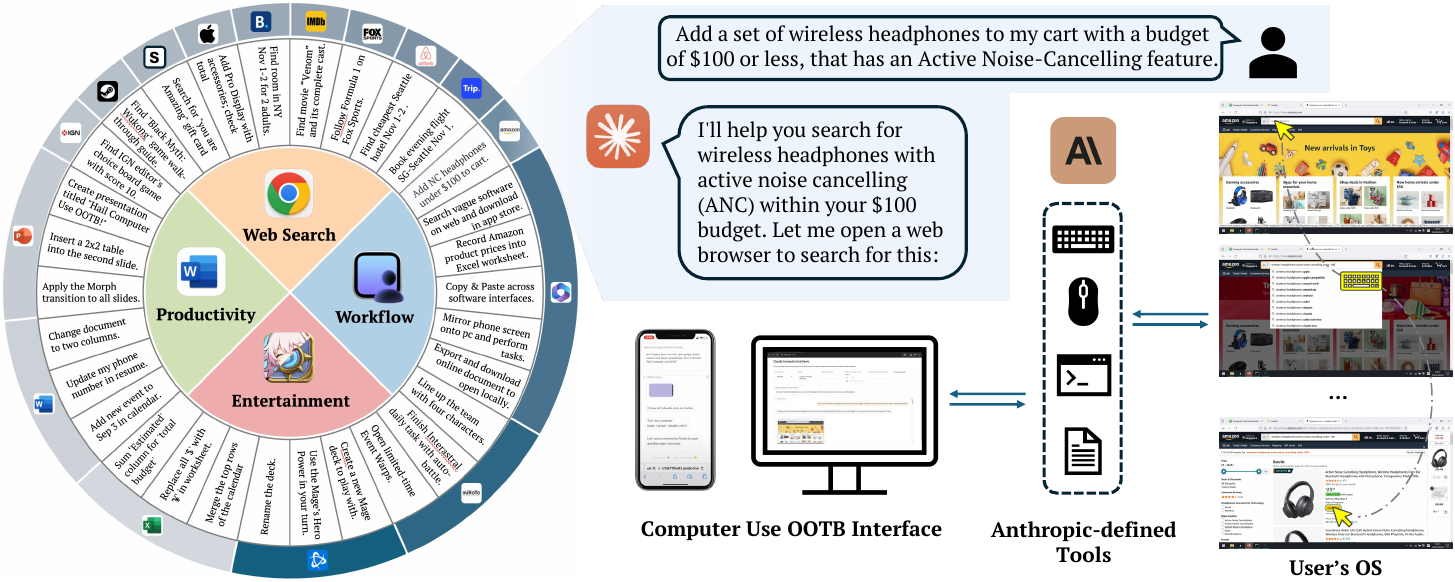}}
\caption[Caption for LOF]{
    Overview of representative evaluation tasks (left), categorized by Web Search, Productivity, Workflow, and Entertainment. Our Computer Use Out-of-the-Box framework (right) provides an easy implementation to execute these tasks in the user's OS.}
\label{fig:example_mainfig}
% \vspace{-0.1in}
\end{figure*}

\section{Introduction}

Automating desktop tasks has become an increasingly popular area of research, driven by the need to enhance users' productivity and accessibility across various application environments. From web navigation to professional software and even video games, users frequently encounter repetitive tasks that could benefit from automation. While large language models like GPT-4 and Qwen-2-VL have demonstrated their potential in automating tasks through general GUI interaction, the capacity of these models is still far from enough for applicable desktop task automation.

Recent studies in GUI automation agents have leveraged general-purpose LLMs to interact with graphical user interfaces (GUIs) by understanding the GUI state and generating actions. However, the release of \textbf{Claude 3.5 Computer Use} by Anthropic marks a significant advancement in this domain, introducing the first frontier AI model to offer computer use in public beta. Unlike previous models, \textbf{Claude 3.5 Computer Use} offers an end-to-end solution through API calls, actions will be generated from user instruction and observed purely visual GUI state, without requiring further external knowledge such as reference plan and GUI parsing.  

Despite this advancement, the community needs a comprehensive analysis that evaluates the performance of API-based GUI automation models in depth. 
To take the first steps to explore the capacities and limitations of such models, we propose a comprehensive case study based on real-world desktop environments, encompassing a diverse range of software domains, including web navigation, professional tools, and games. The selected cases are designed to reflect the needs of various user groups, ensuring that the evaluation covers a broad spectrum of desktop automation tasks.

To isolate specific aspects of the model's capability, we evaluate the performance of API-based GUI automation models rigorously across three dimensions:
\vspace{-4pt}
% \vspace{-3pt}
\begin{itemize}[leftmargin=25pt]
\setlength\itemsep{-0.5pt}
    \item \textbf{Planning}: Assessing the model’s ability to generate an executable plan from the user’s query. The plan should have a correct flow, allowing the overall successful operations of the software, with each step being clear and executable.
    \item \textbf{Action}: Evaluating whether the model can accurately ground the interactable GUI elements and execute the action step-by-step from the derived plan.
    \item \textbf{Critic}: Measuring the model’s awareness of the changing environment, including its ability to adapt to the outcomes of its actions, such as retrying tasks if unsuccessful or terminating execution when the task is completed.
\end{itemize}

To our best knowledge, this is the first comprehensive case study on \textbf{Claude 3.5 Computer Use} and API-based GUI automation models. We hope that our research provides the community with valuable insights into the capacities and limitations of these models. Our case study aim to lay the foundation for the continued exploration and benchmarking of API-based GUI automation. Additionally, to facilitate the community to discover and benchmark the newly released model, we also release an out-of-the-box universal framework, namely \href{https://github.com/showlab/computer_use_ootb}{Computer Use OOTB}, providing a seamless solution for users and researchers to deploy these models in local environments without the need for complex setup or configuration, aiming to improve the accessibility of GUI automation research field.

Our contributions in this report are summarized as follows.

\vspace{-4pt}
% \vspace{-3pt}
\begin{itemize}[leftmargin=25pt]
\setlength\itemsep{-0.5pt}
    \item We present a comprehensive case study for \textbf{Claude 3.5 Computer Use} on desktop task automation, covering domains such as web search, professional software, and games, designed to reflect the needs of various user groups.
    \item We introduce an out-of-the-box, cross-platform agent framework for deploying API-based GUI automation models, offering a universal solution for easy implementation and benchmarking.
    \item We conduct extensive human evaluations and provide in-depth analyses, demonstrating both the advancements and limitations of the newly released API-based GUI automation model.
\end{itemize}

%% file: sections/2-related_work.tex
\section{Related Work}

\paragraph{Large Vision-Language Models}
Recent research has invested tremendous effort in constructing LVLMs capable of jointly processing image and text \citep{liu2023llava, zhu2023minigpt, ye2023mplug, li2023otter},
integrating vision encoders with LLMs through connecting layers,
% LVLMs integrate vision encoders like ViT \citep{dosovitskiy2020image}, with LLMs such as LLaMA \citep{touvron2023llama}, 
inheriting LLMs' linguistic and reasoning skills to perform vision-language tasks.
A series of studies focused on grounding with LVLMs \citep{wang2023visionllm, bai2023qwen, chen2023minigpt}, such as providing bounding boxes for objects when generating responses \citep{chen2023shikra, peng2023kosmos}.

\paragraph{GUI Agents}

Autonomous agents powered by large language models (LLMs), referred to as language agents~\citep{weng2023prompt, sumers2023cognitive}, have gained significant attention due to their interactive capabilities~\citep{renda_survey, sun2023push, hong2024metagpt, durante2024agentaisurvey}. Recent efforts have enabled these agents to interact with operating systems through programs~\citep{sun2024survey} or API calls~\citep{wu2024copilot, zhang2024ufo}. However, the closed-source nature of most commercial software imposes significant limitations, as agents often lack access to internal APIs or code. Consequently, research has shifted toward GUI-based agents that interact with digital devices through human-like mouse and keyboard actions~\citep{cheng2024seeclick, hong2024cogagent, zheng2024seeact}. Models like WebGPT~\cite{nakano2021webgpt}, Agent-Lumos~\cite{yin2024agent}, CogAgent~\cite{hong2024cogagent}, AutoWebGLM~\cite{lai2024autowebglm}, Auto-GUI~\cite{zhang2023you}, AppAgent~\cite{zhang2023appagent}, ScreenAgent~\cite{niu2024screenagent}, and AssistGUI~\cite{gao2023assistgui} have demonstrated improved performance across various tasks, expanding from web navigation to general GUI automation. 

To enhance the effectiveness of these GUI agents, researchers have focused on developing systems that can interpret human intentions and predict actions in the form of function calls~\citep{zhang2024xlam, zhang2024agentohana, zeng2023agenttuning, yin2023lumos}. Nonetheless, progress is hindered by the limited quantity and vast diversity of available agent data~\citep{li2024ac, xu2024envisions}. Specifically, GUI agents remain underexplored, with only a few attempts made to train models that effectively ground GUI interactions~\citep{cheng2024seeclick, hong2024cogagent, gou2024uground}.

Additionally, SearchAgent~\cite{koh2024tree} introduces an inference-time search algorithm to enhance multi-step reasoning and planning in interactive web environments. Collectively, these advancements contribute to the development of more sophisticated and capable GUI agents, pushing the boundaries of automated task completion across various digital platforms.

%% file: sections/3-method.tex
\section{Claude Computer Use Revealed}

% what is ootb? a [ ] framwork?
To establish a robust and in-depth analysis of Claude's Computer Use, we will thoroughly explore the model design and present a framework for the community to replicate. Our analysis will draw on various perspectives, emphasizing both the underlying model and its tools.

\subsection{Model Design}

The main task of Claude Computer Use can be formulated as follows: when presented with a user instruction $X_{instr}$ in natural language, the agent is asked to complete a series of actions on the desktop to complete this instruction. The entire process of agent-environment interactions from initial to final states involves multiple steps. At each time step $t$, the agent will observe the GUI state $I^t$, then decide the next step action from its action space, perform the action with corresponding tools in order to complete the task, afterwards, the model will reflect on the action outcome to enhance its future planning. Following this, we will delve into the detailed design of Claude Computer Use.

% \noindent\textbf{Task description.} To describe the task, a textual request $X_{instr}$ is provided by the user, which describes the functionality of an application to be accomplished, e.g., \emph{Add a set of wireless headphones to your cart with a budget of \$100 or less, that has an active noise-canceling feature}. Afterward, the agent is asked to complete a series of actions on the desktop to complete this instruction. The entire process of agent-environment interactions from initial to final states is called an episode. At each time step $t$ of an episode, the agent will take a screenshot $I^t$, and decide the next step action to take to complete the task.

\subsubsection{System Prompt} Below is the system prompt of Claude Computer Use, where environment-specific variables will be denoted in full capital letters and enclosed in square brackets.

\begin{textblock}[frametitle={\textbf{System Prompt}}]
\begin{lstlisting}
System Overview:
You have access to a set of functions that allow you to interact with a sandboxed computing environment. 
You do NOT have access to external resources, except through the functions provided below.

You can invoke one or more functions by writing a <antml:function_calls> block like this:

<antml:function_calls>
  <antml:invoke name="$FUNCTION_NAME">
    <antml:parameter name="$PARAMETER_NAME">$PARAMETER_VALUE</antml:parameter>
    ...
  </antml:invoke>
  <antml:invoke name="$FUNCTION_NAME2">
    ...
  </antml:invoke>
</antml:function_calls>

String and scalar parameters should be passed as is. Lists and objects should be passed in JSON format.
The output or any errors will appear in a subsequent <function_results> block. You can then respond to the user based on the results or make further function calls.

If a <function_results> block does NOT appear, your function call was likely malformatted.

Available Functions:

1. Computer Interaction (GUI):
   - Description: 
     Use a mouse and keyboard to interact with the computer and take screenshots.
     You can only interact with the desktop GUI (no terminal or application menu access).
   - Actions include:
     - key: Press a key or key-combination.
     - type: Type a string of text.
     - mouse_move: Move the cursor to specified coordinates.
     - left_click, right_click, middle_click, double_click: Perform mouse clicks.
     - left_click_drag: Click and drag the cursor.
     - screenshot: Take a screenshot of the screen.
   - Important Notes:
     - The screen resolution is [SCREEN_RESOLUTION, e.g., 1024x768].
     - Always check the coordinates of elements via screenshots before moving the cursor.
     - If a click fails, adjust your cursor position and retry.
   - Parameters:
     - action (required): The action to perform, such as key, type, mouse_move, etc.
     - coordinate: The (x, y) coordinates for mouse-related actions.
     - text: The text to type or key to press for type and key actions.

2. Bash Shell Commands:
   - Description: Run commands in a bash shell.
   - Parameters:
     - command (required): The bash command to run.
     - restart: If true, restarts the tool.

3. File Editing Tool:
   - Description: View, create, and edit files.
   - Commands:
     - view: Displays a file or lists directory contents.
     - create: Creates a new file (fails if the file already exists).
     - str_replace: Replaces a specific string in a file.
     - insert: Inserts a string after a specified line.
     - undo_edit: Reverts the last edit made to the file.
   - Parameters:
     - path (required): The absolute path to the file or directory.
     - file_text: The content for creating a file.
     - new_str, old_str: Strings for replacing or inserting content.
     - insert_line: Line number for inserting content.
     - view_range: Specify a range of lines to view.

System Capabilities:
You are using an Ubuntu virtual machine with aarch64 architecture.
You can install applications using apt or pip.
Firefox is installed (use the firefox-esr version).
GUI applications can be started from the bash shell using DISPLAY=:1.
The current date is [DATETIME, e.g., Wednesday, October 23, 2024].

Important Notes:
- Firefox Wizard: If the startup wizard appears, ignore it. Do not click "skip this step." Instead, click on the address bar and enter the appropriate URL or search term.
- PDF Handling: If a PDF appears, it may be better to download it using curl and convert it to text using pdftotext for easier reading.

Summary of How to Use the Tools:
- Function Invocation: To interact with the environment, use the <antml:function_calls> block.
- Error Handling: If no <function_results> appear, check for malformatted calls.
- Multiple Calls: Where possible, chain multiple function calls to optimize workflow.
\end{lstlisting}
\end{textblock}

\subsubsection{State Observation} Claude Computer Use observes the environment solely through visual information obtained from real-time screenshots, without relying on metadata or HTML. These screenshots are captured during task operation, enabling the model to effectively imitate human desktop interactions. This capability is crucial for adapting to the highly dynamic nature of the GUI environment. By embracing the "vision-only" approach, Claude Computer Use achieves general computer use without relying on software APIs to perceive the environmental information, particularly for closed-source software.

\subsubsection{Reasoning Paradigm} Claude Computer Use employs a reasoning-acting paradigm for its reasoning process, generating more reliable actions in the highly dynamic GUI environment. Similar to traditional ReAct~\cite{yao2023react}, Claude Computer Use observes the environment before deciding on an action, ensuring that the action is appropriate for the current GUI state. Furthermore, Claude Computer Use exhibits the capacity to efficiently identify when user requirements are fulfilled, enabling it to take decisive actions without engaging in unnecessary steps. Interestingly, beyond traditional ReAct paradigm, which typically involves continuous observation of the environment at each step, Claude Computer Use adopts a more selective observation strategy. It monitors the GUI state only when necessary, according to its reasoning. This approach effectively reduces costs and accelerates the overall process by avoiding superfluous observations.

\subsubsection{Tool Use} Currently, Claude Computer Use is provided with three Anthropic-defined tools: \textbf{Computer Tools}, \textbf{Text Editor Tools}, and \textbf{Bash Tools}. Below are detailed descriptions of each tool:

\noindent\textbf{Computer Tools.} Computer tools help Claude Computer Use operate a mouse and keyboard to interact with a computer, and take screenshots.

Below is the description of Computer Tools:

\begin{itemize}
    \item This is an interface to a desktop GUI. You do not have access to a terminal or applications menu. You must click on desktop icons to start applications.
    \item Some applications may take time to start or process actions, so you may need to wait and take successive screenshots to see the results of your actions. E.g. if you click on Firefox and a window doesn't open, try taking another screenshot.
    \item The screen's resolution is \texttt{{\{display\_width\_px\}}}x\texttt{{\{display\_height\_px\}}}.
    \item The display number is \texttt{{\{display\_number\}}}.
    \item Whenever you intend to move the cursor to click on an element like an icon, you should consult a screenshot to determine the coordinates of the element before moving the cursor.
    \item If you tried clicking on a program or link but it failed to load, even after waiting, try adjusting your cursor position so that the tip of the cursor visually falls on the element that you want to click.
    \item Make sure to click any buttons, links, icons, etc. with the cursor tip in the center of the element. Don't click boxes on their edges unless asked.
\end{itemize}

Below is the tool schema of Computer Tools:

\begin{jsonframe}[frametitle={\textbf{Computer Tool Schema}}]
\begin{lstlisting}[language=json]
{
    "properties": {
        "action": {
            "description": """The action to perform. The available actions are:
                * `key`: Press a key or key-combination on the keyboard.
                  - This supports xdotool's `key` syntax.
                  - Examples: "a", "Return", "alt+Tab", "ctrl+s", "Up", "KP_0" (for the numpad 0 key).
                * `type`: Type a string of text on the keyboard.
                * `cursor_position`: Get the current (x, y) pixel coordinate of the cursor on the screen.
                * `mouse_move`: Move the cursor to a specified (x, y) pixel coordinate on the screen.
                * `left_click`: Click the left mouse button.
                * `left_click_drag`: Click and drag the cursor to a specified (x, y) pixel coordinate on the screen.
                * `right_click`: Click the right mouse button.
                * `middle_click`: Click the middle mouse button.
                * `double_click`: Double-click the left mouse button.
                * `screenshot`: Take a screenshot of the screen.""",
            "enum": [
                "key",
                "type",
                "mouse_move",
                "left_click",
                "left_click_drag",
                "right_click",
                "middle_click",
                "double_click",
                "screenshot",
                "cursor_position"
            ],
            "type": "string"
        },
        "coordinate": {
            "description": "(x, y): The x (pixels from the left edge) and y (pixels from the top edge) coordinates to move the mouse to. Required only by `action=mouse_move` and `action=left_click_drag`.",
            "type": "array"
        },
        "text": {
            "description": "Required only by `action=type` and `action=key`.",
            "type": "string"
        }
    },
    "required": ["action"],
    "type": "object"
}
\end{lstlisting}
\end{jsonframe}

\noindent\textbf{Editor Tools.} Computer tools help Claude Computer Use operate custom editing tool for viewing, creating, and editing files.

Below is the description of Editor Tools:

\begin{itemize}
    \item State is persistent across command calls and discussions with the user.
    \item If \texttt{path} is a file, \texttt{view} displays the result of applying \texttt{cat -n}. If \texttt{path} is a directory, \texttt{view} lists non-hidden files and directories up to 2 levels deep.
    \item The \texttt{create} command cannot be used if the specified \texttt{path} already exists as a file.
    \item If a \texttt{command} generates a long output, it will be truncated and marked with \texttt{<response clipped>}.
    \item The \texttt{undo\_edit} command will revert the last edit made to the file at \texttt{path}.
\end{itemize}

\begin{itemize}
    \item[] \textbf{Notes for using the \texttt{str\_replace} command:}
    \item The \texttt{old\_str} parameter should match EXACTLY one or more consecutive lines from the original file. Be mindful of whitespaces!
    \item If the \texttt{old\_str} parameter is not unique in the file, the replacement will not be performed. Make sure to include enough context in \texttt{old\_str} to make it unique.
    \item The \texttt{new\_str} parameter should contain the edited lines that should replace the \texttt{old\_str}.
\end{itemize}

Below is the tool schema of Editor Tools:

% Use the jsonframe environment with the listings package for syntax highlighting
\begin{jsonframe}[frametitle={\textbf{Editor Tool Schema}}]
\begin{lstlisting}[language=json]
{
    "properties": {
        "command": {
            "description": "The commands to run. Allowed options are: `view`, `create`, `str_replace`, `insert`, `undo_edit`.",
            "enum": ["view", "create", "str_replace", "insert", "undo_edit"],
            "type": "string"
        },
        "file_text": {
            "description": "Required parameter of `create` command, with the content of the file to be created.",
            "type": "string"
        },
        "insert_line": {
            "description": "Required parameter of `insert` command. The `new_str` will be inserted AFTER the line `insert_line` of `path`.",
            "type": "integer"
        },
        "new_str": {
            "description": "Optional parameter of `str_replace` command containing the new string (if not given, no string will be added). Required parameter of `insert` command containing the string to insert.",
            "type": "string"
        },
        "old_str": {
            "description": "Required parameter of `str_replace` command containing the string in `path` to replace.",
            "type": "string"
        },
        "path": {
            "description": "Absolute path to file or directory, e.g. `/repo/file.py` or `/repo`.",
            "type": "string"
        },
        "view_range": {
            "description": "Optional parameter of `view` command when `path` points to a file. If none is given, the full file is shown. If provided, the file will be shown in the indicated line number range, e.g. [11, 12] will show lines 11 and 12. Indexing at 1 to start. Setting `[start_line, -1]` shows all lines from `start_line` to the end of the file.",
            "items": {"type": "integer"},
            "type": "array"
        }
    },
    "required": ["command", "path"],
    "type": "object"
}
\end{lstlisting}
\end{jsonframe}

\noindent\textbf{Bash Tools.} Computer tools help Claude Computer Use run commands in a bash shell.

Below is the description of Bash Tools:

\begin{itemize}
    \item When invoking this tool, the contents of the \texttt{command} parameter does NOT need to be XML-escaped.
    \item You have access to a mirror of common Linux and Python packages via \texttt{apt} and \texttt{pip}.
    \item State is persistent across command calls and discussions with the user.
    \item To inspect a particular line range of a file, e.g. lines 10-25, try \texttt{sed -n 10,25p /path/to/the/file}.
    \item Please avoid commands that may produce a very large amount of output.
    \item Please run long-lived commands in the background, e.g. \texttt{sleep 10 \&} or start a server in the background.
\end{itemize}

Below is the tool schema of Bash Tools:

\begin{jsonframe}[frametitle={\textbf{Bash Tool Schema}}]
\begin{lstlisting}[language=json]
{
    "properties": {
        "command": {
            "description": "The bash command to run. Required unless the tool is being restarted.",
            "type": "string"
        },
        "restart": {
            "description": "Specifying true will restart this tool. Otherwise, leave this unspecified.",
            "type": "boolean"
        }
    }
}
\end{lstlisting}
\end{jsonframe}

\subsubsection{GUI Action space} The GUI action space of Claude Computer Use consists of all the raw mouse and keyboard actions, including mouse-move, left-click, right-click, middle-click, double-click, drag, type, keystrokes, and combinations of keys for shortcuts, among others. Coordinate-related operations also include the target position at the pixel space of the observed screenshot. Therefore, one action can denoted by the syntax \texttt{action\_type(arguments)}. Here are some examples of actions that are supported in our case study:

% \texttt{dragTo(100, 100)}, which indicates the execution of a drag action from the current position to the coordinate (100, 100).
\begin{itemize}[leftmargin=25pt, itemsep=0pt]
    \item \textbf{Mouse Movement:}
    {\setlength{\parskip}{0pt} Move the mouse cursor to a specific position on the screen.\\
    \textit{Example:} \texttt{mouse\_move(100, 150)}}
    
    \item \textbf{Mouse Clicks:}
    {\setlength{\parskip}{0pt} Perform mouse clicks at a specified location.\\
    \textit{Example:} \texttt{left\_click()}}
    
    \item \textbf{Typing and Sending Keystrokes:}
    {\setlength{\parskip}{0pt} Simulate typing text or pressing keys.\\
    \textit{Example:} \texttt{type('Hello, world!')}}
    
    \item \textbf{Keyboard Hotkey Combinations:}
    {\setlength{\parskip}{0pt} Press and release keyboard shortcuts or hotkeys.\\
    \textit{Example:} \texttt{key('ctrl + c')}}
    
    \item \textbf{Drag and Drop:}
    {\setlength{\parskip}{0pt} Perform drag and drop actions.\\
    \textit{Example:} \texttt{left\_click\_drag(100, 200, duration=2)}}

    \item \textbf{Taking Screenshots:}
    {\setlength{\parskip}{0pt} Taking screenshots from the computer to observe.\\
    \textit{Example:} \texttt{screenshot()}}
    
\end{itemize}

\subsubsection{History Visual Context Maintenance} Claude Computer Use maintains an extensive context of history screenshots, which accumulate through the ongoing task operations. Specifically, at each time step, the retained screenshots are utilized to assist the action generation process as follows:
\begin{align}
    Y^t_{action} &= \Theta_{model}(X_{instr}, I^t, I^{t-1}_{history}) \\
    Y^{t-1}_{history} &= I^{t-1} \cup I^{t-2}_{history}
\end{align}
where \(Y^t_{action}\) is the action to take at the current step \(t\), and \(I^{t-1}_{history}\) represents the retained historical screenshots. Here, \(\Theta_{model}\) is the parameterized Claude 3.5 Sonnet model. In this way, the full visual information along the trace of history is preserved, enhancing the model's ability to make informed decisions as an episode unfolds.

%in the end. 
% We identify two main challenges:~(1) 

% \subsection{Interacting with \gptname}
% Prompt Engineering has been an emergent topic for natural language processing and large language models~\citep{wei2022chain,kojima2022large}. With the development of recent instruction-following models~\citep{openai2023gpt4,touvron2023llama}, language models have been less dependant on the prompts. In this paper, we are mainly interested in exploring how to use \gptname for smartphone navigation and designing methods to enable this function, and do not focus on finding optimal prompt designs. Nevertheless, we performed ablation studies and robustness check on different prompt templates we used, as shown in~\cref{sec:android_ablation}.

\subsection{Agent Implementation}

\subsubsection{Out-of-the-Box Agent Framework}
Recognizing that the demonstration codebase from Anthropic only supports a Docker Linux environment, which is far from enough for benchmarking GUI automation models in real-world environments, we have developed a cross-platform, Docker-free GUI agent framework called \href{https://github.com/showlab/computer_use_ootb}{\textbf{Computer Use Out-of-the-Box}}. This framework enables the deployment of a GUI agent locally on both Windows and macOS. By utilizing PyAutoGUI, we ensure that the operations are compatible across both operating systems, allowing universal remote control of the software by the API-based model through specific action commands.

%% file: sections/4-CUAE.tex
\begingroup
\definecolor{tablelinkcolor}{RGB}{203, 93, 82}
\hypersetup{linkcolor=tablelinkcolor} % PlanningColor
\begin{table}[!h]
\centering
\addtolength{\leftskip} {-2cm}
\addtolength{\rightskip}{-2cm}
\renewcommand{\arraystretch}{1.25} % Increase row height by 1.5x
\caption{Summary of case studies in the report. Click on tasks to navigate to corresponding sections.}
\vspace{0.05cm}
\label{tab:case_studies_summary}
% \begin{tabular}{|l|l|p{8.5cm}|c|}
\begin{tabular}{|l|l|l|c|}
\hline
\textbf{Domain} & \textbf{Site / Software} & \textbf{Task} & \textbf{Outcome} \\ \hline\hline
Web Search & Amazon & \hyperref[sec:Find ANC]{Find ANC Headphones Under Budget \$100 on Amazon} & Success \\ \hline
Web Search & Apple Official Site& \hyperref[sec:Browse Apple]{Browse Apple Official Site for Display with Accessories} & Success \\ \hline
Web Search & Fox Sport & \hyperref[sec:Fox Sports]{Fox Sports Subscription} & Failed \\ \hline
Workflow & Apple Music & \hyperref[sec:Find Music]{Find Latest \& Local Trending Music and Add to Playlist} & Success \\ \hline
Workflow & Amazon \& Excel & \hyperref[sec:Search for Products and Record]{Search for Products on Amazon and Record Prices in Excel} & Success \\ \hline
Workflow & Google Sheet \& Excel & \hyperref[sec:Export and Download]{Export and Download Online Document to Open Locally} & Success \\ \hline
Workflow & App Store & \hyperref[sec:Install App]{Install App from App Store and Report Storage Usage} & Success \\ \hline
Office Productivity & Outlook & \hyperref[sec:Forward Email]{Forward a Specific Email and CC Another Recipient} & Success \\ \hline
Office Productivity & Word & \hyperref[sec:Change Document Layout to A3]{Change Document Layout to A3 in Landscape Orientation} & Success \\ \hline
Office Productivity & Word & \hyperref[sec:Two Columns]{Two Columns Document} & Success \\ \hline
Office Productivity & Word & \hyperref[sec:Update Name and Phone]{Update Name and Phone Number on Resume Template} & Failed \\ \hline
Office Productivity & PowerPoint & \hyperref[sec:Gradient Fill]{Gradient Fill Background} & Success \\ \hline
Office Productivity & PowerPoint & \hyperref[sec:Modify Slide Title]{Modify Slide Title and Draw a Triangle} & Success \\ \hline
Office Productivity & PowerPoint & \hyperref[sec:Insert Numbering]{Insert Numbering Symbol} & Failed \\ \hline
Office Productivity & Excel & \hyperref[sec:Find and Replacement]{Find and Replacement in Worksheet} & Success \\ \hline
Office Productivity & Excel & \hyperref[sec:Insert a Sum]{Insert a Sum Equation over Cells} & Failed \\ \hline
Video Games & \textit{Hearthstone} & \hyperref[sec:Create and Rename a New Deck]{Create and Rename a New Deck for Battle} & Success \\ \hline
Video Games & \textit{Hearthstone} & \hyperref[sec:Hero Power]{Hero Power} & Success \\ \hline
Video Games & \textit{Honkai: Star Rail} & \hyperref[sec:Warp]{Warp Automation} & Success \\ \hline
Video Games & \textit{Honkai: Star Rail} & \hyperref[sec:Daily Mission Clean up]{Daily Mission Clean up Automation} & Success \\ \hline
\end{tabular}
\renewcommand{\arraystretch}{1} % Reset row height back to normal for future tables
% \vspace{-0.2cm}
\end{table}
\endgroup

\section{Computer Use Ability Evaluation}

\subsection{Setup Details}

\paragraph{System Config.}
The evaluation is conducted on both Windows and macOS via the proposed \textbf{Computer Use Out-of-the-Box} platform. As suggested by \href{https://github.com/anthropics/anthropic-quickstarts/tree/main/computer-use-demo#screen-size}{Anthropic Computer use API document} \citep{anthropic2024claude35}, the resolution is set to (1366, 768) and (1344, 756) for Windows and macOS, respectively.

\paragraph{Human Review and Evaluation.}
Computer use introduces extra risks that differ significantly from those standard conversational APIs or interfaces, especially when interacting with the internet, or potentially manipulating users' sensitive information. Thus, we use a human evaluation to continuously monitor and review the process. We also manually observe the final state of a task upon completion and determine outcomes as a "Success" or "Failed".

\paragraph{Case Study Scope.}
As shown in Figure \ref{fig:example_mainfig} (left), we carefully collected a set of user queries and initial states on the following widely-applicated domains to include a broad spectrum of desktop tasks across operating systems. Specifically, in this report, we include 20 tasks across 12 software or websites in the following 3 domains: Web Search, Workflow, Office Productivity and Video Games.

% \subsection{Summary of Case Studies}

Table \ref{tab:case_studies_summary} documents an overview of the case studies evaluated in this section, categorized by domain and indicating whether each task was successfully completed or failed. For quick navigation, we suggest readers to click on and navigate to their scenarios of interest via \hyperref[tab:case_studies_summary]{this table}.

\subsection{Case Study: Web Search}

The World Wide Web (WWW) is a vast, open-domain interactive environment consisting of interconnected pages with natural text, images, and numerous interactive elements. The dynamic nature of web pages means that web search tasks place significant demands on a model's planning capabilities, as the model cannot simply follow a pre-defined path. Additionally, the large number of interactive elements requires robust grounding abilities to identify and interact with the correct elements accurately. Furthermore, given the interconnected structure of web pages, the model should be capable of deciding when to navigate forward or backward through the history trace, depending on the execution status of each planned step. Therefore, the model must possess the ability to critique and adjust its plan accordingly.

In the subsequent case studies, we evaluate the model's performance on complex web search tasks that reflect real-world scenarios. These tasks are designed to assess the model's planning, grounding, and adaptive abilities when navigating through intricate web interfaces. By examining how the model interacts with actual websites, we aim to demonstrate its proficiency in handling dynamic content, executing multi-step plans, and adjusting its strategies in response to unexpected interface behaviors. This evaluation provides insights into both the capabilities and limitations of the model within the web search domain.

\begin{figure*}[!h]
\centering
\centerline{\includegraphics[width=1.2\linewidth]{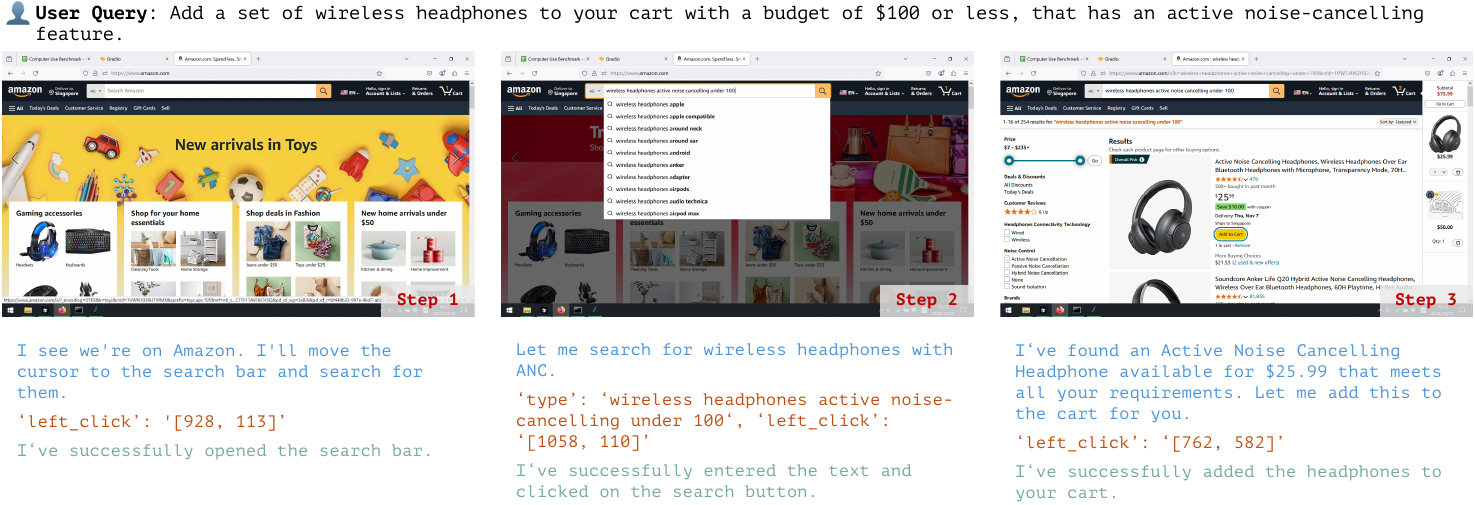}}
\caption[Caption for LOF]{
Representative task in Web Search. We categorize and highlight the response of the model into: \textcolor{PlanningColor}{{\fontfamily{lmtt}\selectfont \textbf{Planning}}}, \textcolor{ActionColor}{{\fontfamily{lmtt}\selectfont \textbf{Action}}}, and \textcolor{CriticColor}{{\fontfamily{lmtt}\selectfont \textbf{Critic}}}. Please check Section \ref{sec:Find ANC} for detailed discussions. Zoom in for the best view.
	}
\label{fig:example_web2}
\vspace{-0.1in}
\end{figure*}

\subsubsection{Find ANC Headphones Under Budget \$100 on Amazon} \label{sec:Find ANC}

\noindent\textbf{\textcolor{PlanningColor}{Planning.}} The blue parts in Figure \ref{fig:example_web2} demonstrate the planning process of the model. It can be observed that the model is aware of its initial status that it is on the Amazon main page, and then comes up with the plan that search for the target headphone using a query extracted from the user instruction through the search box, and filter the result items based on the budget conditions. 

\noindent\textbf{\textcolor{ActionColor}{Action.}} The brown parts in the figures demonstrate the action generated from the model. It can be observed that the model can accurately locate the center of the search box and perform click operation. After typing in the search box, it can further click on the search logo to start the search. 

\noindent\textbf{\textcolor{CriticColor}{Critic.}} It can be observed that after Claude clicks on the "Add to Cart" button, the model takes another screenshot to check it has successfully added the desired item to the cart. This indicates that the model can observe the outcome of its action to dynamically decide whether to retry or terminate the execution.

% TODO 这一页要写满

% \newpage

\begin{figure*}[!h]
\vspace{-0.2in}
\centering
\centerline{\includegraphics[width=1.18\linewidth]{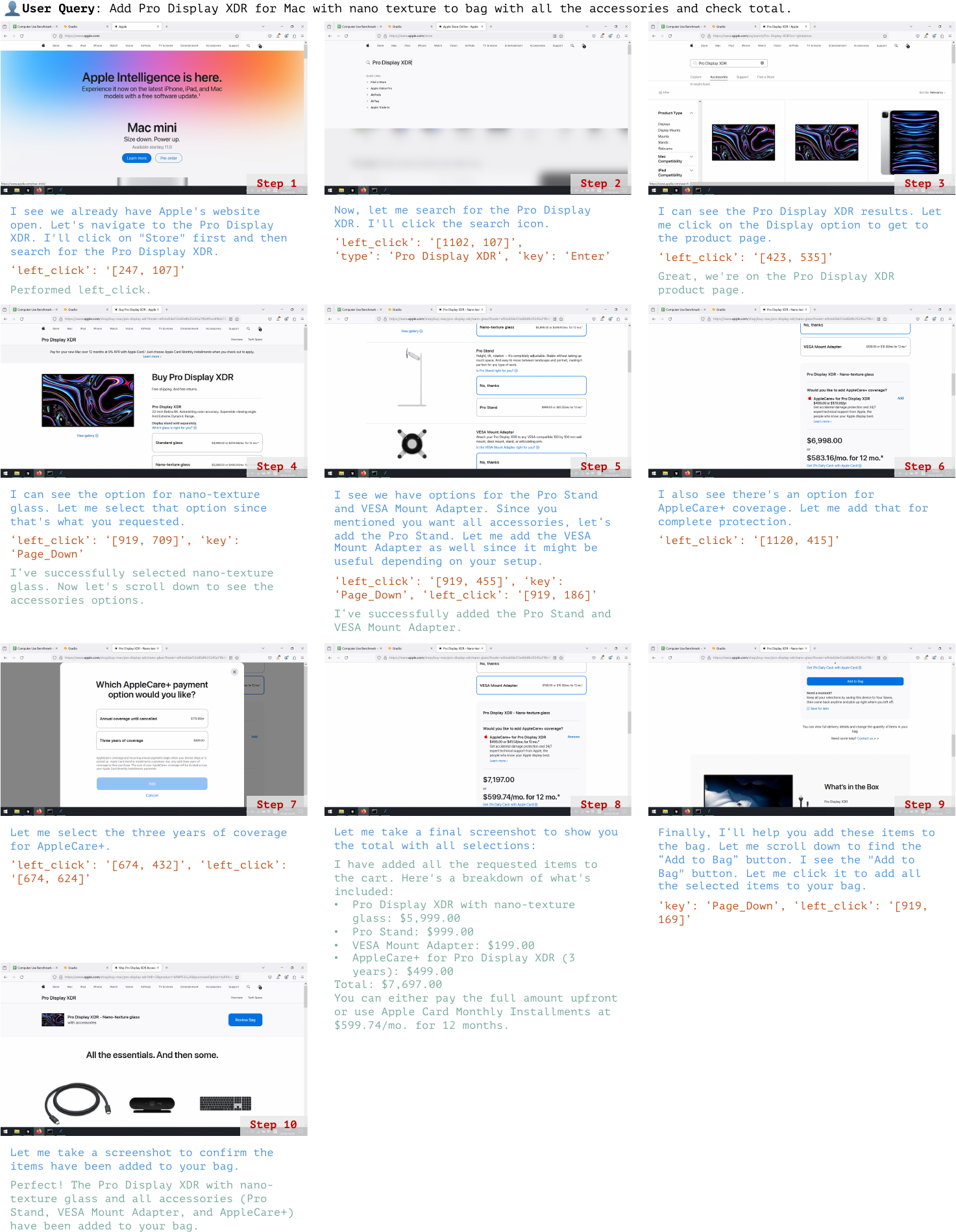}}
\caption[Caption for LOF]{
Representative task in Web Search. We categorize and highlight the response of the model into: \textcolor{PlanningColor}{{\fontfamily{lmtt}\selectfont \textbf{Planning}}}, \textcolor{ActionColor}{{\fontfamily{lmtt}\selectfont \textbf{Action}}}, and \textcolor{CriticColor}{{\fontfamily{lmtt}\selectfont \textbf{Critic}}}. Please check Section \ref{sec:Browse Apple} for more discussions. Zoom in for the best view.
}
\label{fig:example_web1}
\vspace{-0.1in}
\end{figure*}

% \newpage

\subsubsection{Browse Apple Official Site for Display with Accessories}  \label{sec:Browse Apple}

\noindent\textbf{\textcolor{PlanningColor}{Planning.}} The blue parts in Figure \ref{fig:example_web1} demonstrate the planning process of the model. It can be observed that the model can observe its initial status on the main page of Apple's official site. It is worth noting that if the model plans to find the desired item by navigating through menus and sub-menus, it will be time-consuming and require multiple difficult operations such as cursor hovering and scrolling. Therefore, the model plans to find the target item by utilizing the search function, showcasing the efficiency of the plan generated by the model. Moreover, when the Apple Care window pops up, the model captures the change, and further planning according to the user instruction and options in the pop-up window: Since the user wants all accessories, thus it is reasonable to add the three-year AppleCare+ which has a longer coverage to the cart.  

\noindent\textbf{\textcolor{ActionColor}{Action.}} As shown in the figure, the model interacts with different types of elements, including text, buttons, and even hyperlinked images. This demonstrates the strong grounding capacity of Claude Computer Use, as it utilizes purely visual information only, without HTML metadata.

\noindent\textbf{\textcolor{CriticColor}{Critic.}} Claude Computer Use has demonstrated its strong critic ability through its reflection after key steps, as shown in the green parts. It can be observed that the model determines that it has successfully added the nano-texture option, and continues to the tasks left. Moreover, after all the accessories have been added to the cart, Claude Computer Use makes a final screenshot to perform a final check and list all the accessories it has added. These critical observations greatly improve the model's ability to complete long and dynamic tasks, as in this case.

\subsubsection{Fox Sports Subscription (Failed)} \label{sec:Fox Sports}

\begin{figure*}[!h]
\centering
\centerline{\includegraphics[width=1.2\linewidth]{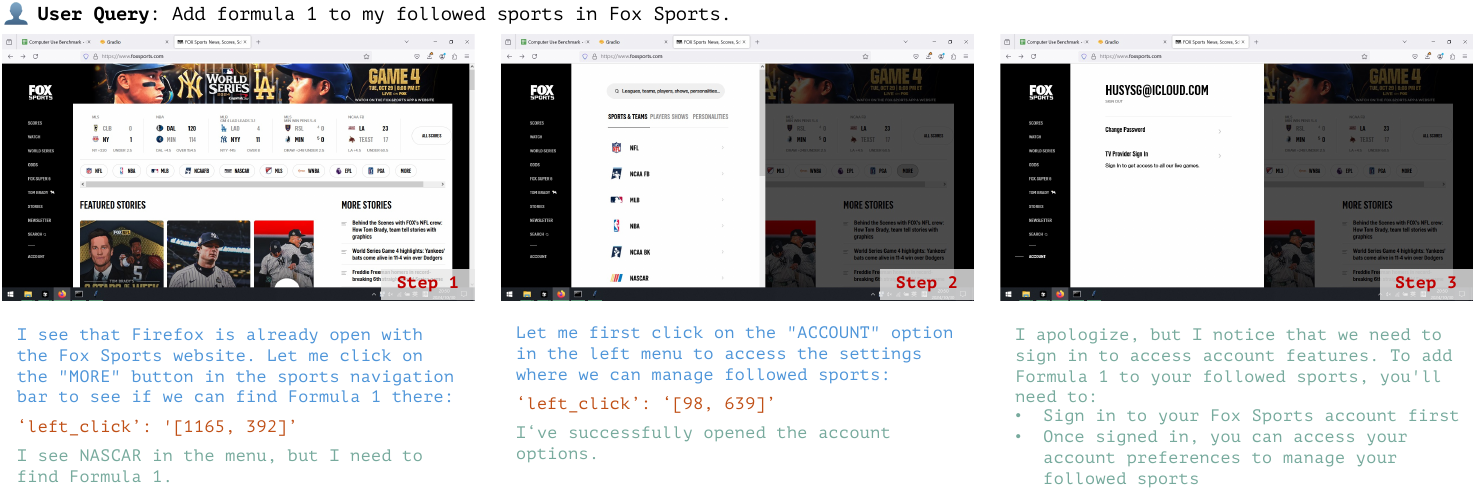}}
\caption[Caption for LOF]{
Representative \textbf{failure} task in Web Search. We categorize and highlight the response of the model into: \textcolor{PlanningColor}{{\fontfamily{lmtt}\selectfont \textbf{Planning}}}, \textcolor{ActionColor}{{\fontfamily{lmtt}\selectfont \textbf{Action}}}, and \textcolor{CriticColor}{{\fontfamily{lmtt}\selectfont \textbf{Critic}}}. 
The 
\textcolor{ErrorColor}{{\fontfamily{lmtt}\selectfont \textbf{Error}}} here is as follows:
\textcolor{ErrorColor}{
\textbf{Scrolling down} the navigation panel is required to find Formula 1 in \textbf{Step 2}. However, the model gives incorrect plan that go to the “Account” tab.
}
Please check Section \ref{sec:Fox Sports} for more discussions. Zoom in for the best view.
}
\label{fig:example_web3}
\vspace{-0.1in}
\end{figure*}

\noindent\textbf{\textcolor{PlanningColor}{Planning.}} 
The blue captions in Figure \ref{fig:example_web3} represent the model’s planning process. In this case, the model recognizes Fox Sports website and decides to look for Formula 1 within the available sports categories. It initially formulates a plan to explore the sports options by selecting the "MORE" button in the navigation menu for more sports categories. When the model does not immediately find Formula 1 in the initial sports list, it alters its approach, deciding to access the "ACCOUNT" menu, with the expectation that this section might allow the user to manage more following sports. 

\noindent\textbf{\textcolor{ActionColor}{Action.}}
In Step 1, the model accurately identifies the location of the "MORE" button in the sports navigation panel and clicks on it, which should expand the list to show additional sports. In Step 2, after failing to find Formula 1 through this initial method, the model adapts its strategy and proceeds to click on the "ACCOUNT" tab in the left-side menu. This transition reflects the model's flexibility in following alternative paths to achieve the user’s intended outcome. The sequence of actions demonstrates the model’s ability to interact with multiple sections of the interface as it attempts to locate the desired content.

\noindent\textbf{\textcolor{CriticColor}{Critic.}} 
The green captions depict the model’s feedback and self-assessment process following its actions. After accessing the "MORE" tab, the model identifies one of other related site and re-emphasizes its targeting sport. Although the final result incorrect, this critic sequence still reflects the model’s attempt to achieve the user’s goal by exploring both direct navigation and re-planning alternative routes.
 This critic phase demonstrates the model’s capacity to adjust its instructions dynamically based on the current interface requirements, also shows its situational awareness when faced with authentication barriers.

\noindent\textbf{\textcolor{ErrorColor}{Error.}} 
The error, highlighted in red in the caption, reveals a significant oversight in the model’s planning. The model initially attempts to locate Formula 1 within the expanded sports categories under the "MORE" button but does not succeed. Instead of continuing to explore the navigation panel through scrolling, the model erroneously shifts its strategy to the "ACCOUNT" tab, mistakenly assuming that account settings might provide the desired sport. This results in an unnecessary detour, as accessing the "ACCOUNT" tab prompts a login requirement, that ultimately misleads task completion and adds unnecessary complexity for the user.

This error highlights the importance of contextually aware navigation, particularly when the model fails to locate an item on the initial interface view — a very common real-world scenario. Instead of prematurely altering its plan, the model should prioritize further scrolling within navigation panels to continue its search. Although this is a brief task, it provides insights into the model’s limitations with scrolling-based navigation, and underscores areas for enhancement. Specifically, refining the model's approach on visual search while maintaining continuity within the interface may strengthen its performance in future versions.

\subsection{Case Study: Workflow}

Workflow tasks involve multi-application interactions or multi-targeting user queries that require the model to seamlessly navigate and manage data across different software platforms. In real-world scenarios, users often need to coordinate actions between web browsers, productivity tools, and specialized applications to accomplish complex objectives. These tasks test the model's ability to maintain context across different environments, handle data transfer accurately, and execute multi-step processes without losing track in intermediate states.

The complexity of workflow tasks lies in the need for robust planning capabilities to coordinate operations across diverse interfaces. The model must possess strong grounding abilities to interpret and interact with varying user interface elements, which may significantly differ or similar between applications. Additionally, it should be capable of dynamically adjusting its plan in response to unexpected behaviors or errors that may occur when switching contexts. In the following case studies, we assess the model's performance on complex workflow tasks that reflect some practical use cases. By examining how the model integrates actions and maintains consistency in multi-application environments or multi-targeting tasks, we aim to demonstrate its capabilities, and identify areas for improvement within the workflow domain.

\subsubsection{Find Latest \& Local Trending Music and Add to Playlist} \label{sec:Find Music}

\begin{figure*}[!h]
\centering
\centerline{\includegraphics[width=1.2\linewidth]{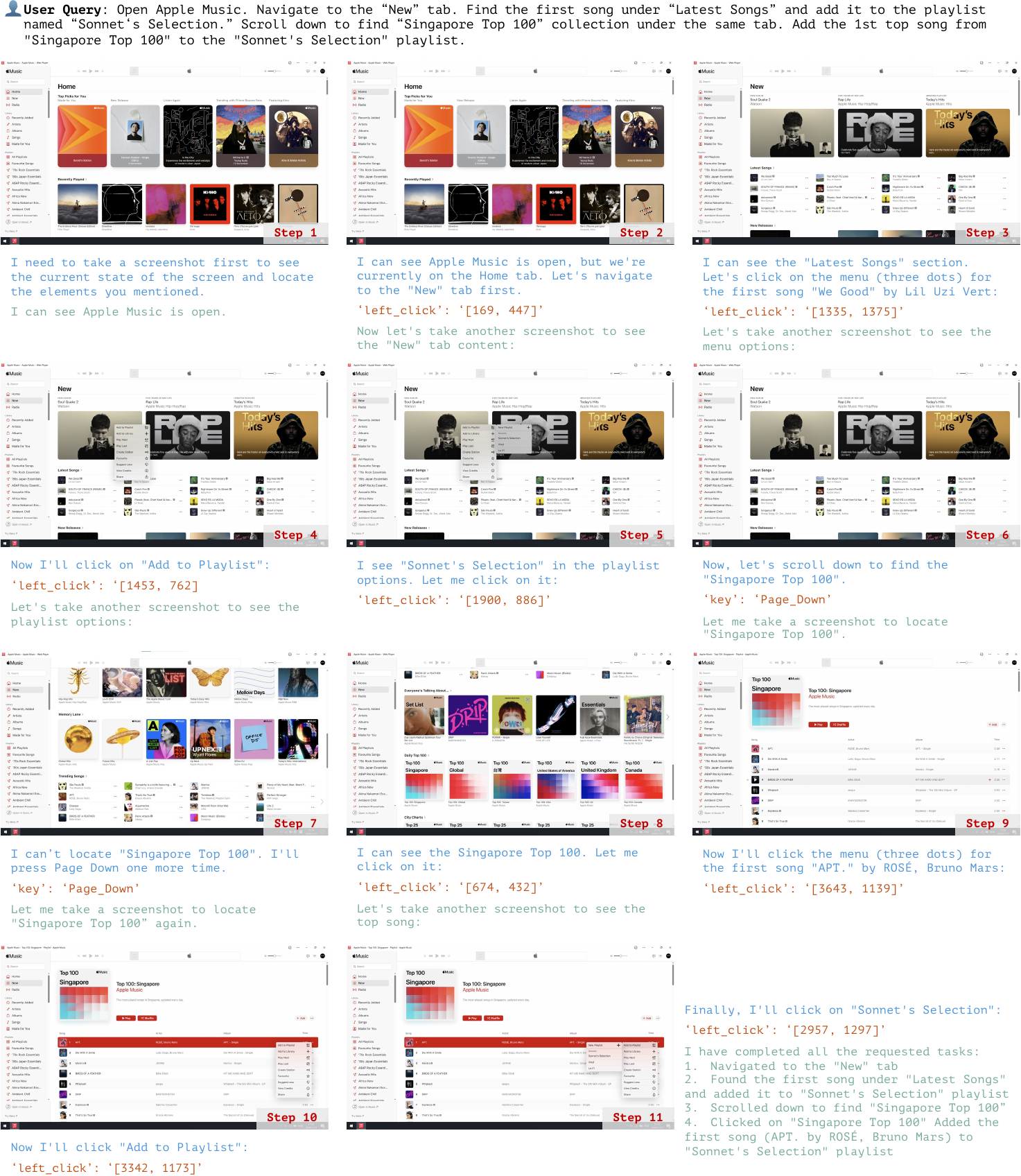}}
\caption[Caption for LOF]{
Representative task in Workflow. We categorize and highlight the response of the model into: \textcolor{PlanningColor}{{\fontfamily{lmtt}\selectfont \textbf{Planning}}}, \textcolor{ActionColor}{{\fontfamily{lmtt}\selectfont \textbf{Action}}}, and \textcolor{CriticColor}{{\fontfamily{lmtt}\selectfont \textbf{Critic}}}. Please check Section \ref{sec:Find Music} for more discussions. Zoom in for the best view.
}
\label{fig:example_wf1}
\vspace{-0.1in}
\end{figure*}

\noindent\textbf{\textcolor{PlanningColor}{Planning.}} 
The blue captions in Figure \ref{fig:example_wf1} illustrate the model’s planning sequence for locating trending music and adding specific songs to a designated playlist within Apple Music. Initially, the model recognizes that it needs to locate the "New" tab within Apple Music to begin the search. Once in the "New" tab, the model plans to find the first song listed under the "Latest Songs" section and to add this song to a pre-existing "Sonnet's Selection" playlist. Following this, the model continues the secondary plan to scroll through the "New" tab to locate the "Singapore Top 100" collection, where it will select the top song and similarly add it to "Sonnet's Selection". This planning phase demonstrates the model’s understanding of a multi-step objective involving tab navigation, section identification, and song selection for playlist addition, guided by the user’s instructions.

\noindent\textbf{\textcolor{ActionColor}{Action.}}
In Step 2, the model initiates navigation by clicking on the "New" tab to transition away from the Home tab. After reaching the "New" tab, the model proceeds to locate the first song under "Latest Songs" and selects the song options menu by clicking on the three-dot icon next to it. Through this menu, the model selects "Add to Playlist" and, upon the display of playlist options, identifies and clicks on "Sonnet's Selection" to add the song as instructed. Following this, the model initiates the second part of the task by scrolling down to locate "Singapore Top 100." The model prefers to use the Page Down key to simulate navigation, with repeatedly taking screenshots to verify its position. Upon locating the "Singapore Top 100" section, the model identifies the first song and repeats the same adding process.

\noindent\textbf{\textcolor{CriticColor}{Critic.}} 
After navigating to the "New" tab, the model confirms that it has successfully reached the appropriate section before proceeding to locate "Latest Songs." This verification process is repeated upon opening the options menu for the first song and after accessing the playlist options. When adding songs to "Sonnet's Selection," the model confirms each action to ensure it aligns with the user’s specified goal. Besides, in the scrolling phase, the model validates each step by periodically capturing screenshots to assess its current position. This iterative verification is essential for ensuring the target is located without a miss. Once the model reaches the top song in "Singapore Top 100" and adds it to the playlist, it provides a final confirmation of task completion. This feedback loop illustrates the model’s capacity for continuous monitoring through repeatedly visual confirmation, which can be critical for tasks involving multi-paged navigation.

\subsubsection{Search for Products on Amazon and Record Prices in Excel} \label{sec:Search for Products and Record}

\begin{figure*}[!h]
\centering
\centerline{\includegraphics[width=1.2\linewidth]{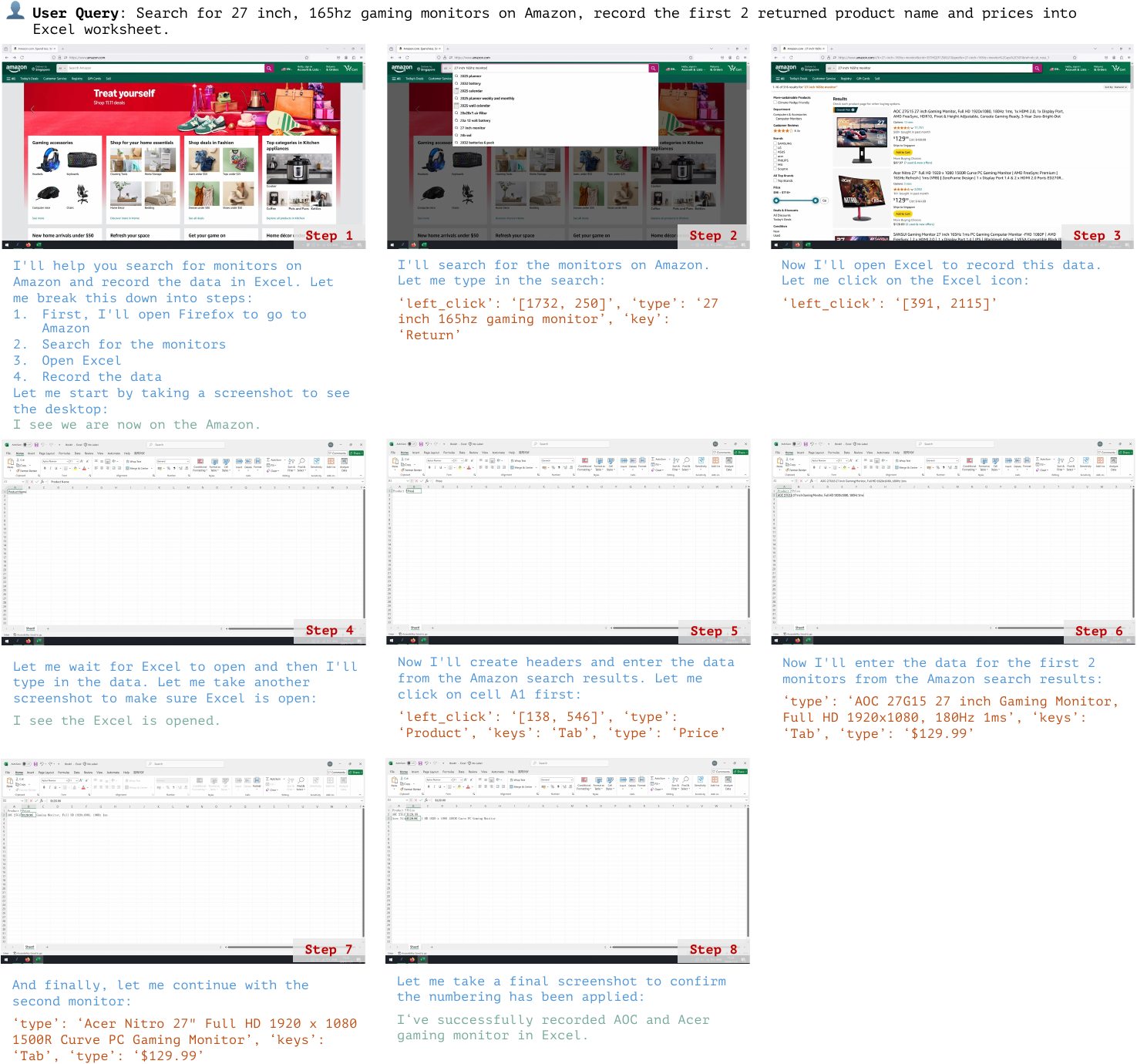}}
\caption[Caption for LOF]{
Representative task in Workflow. We categorize and highlight the response of the model into: \textcolor{PlanningColor}{{\fontfamily{lmtt}\selectfont \textbf{Planning}}}, \textcolor{ActionColor}{{\fontfamily{lmtt}\selectfont \textbf{Action}}}, and \textcolor{CriticColor}{{\fontfamily{lmtt}\selectfont \textbf{Critic}}}. Please check Section \ref{sec:Search for Products and Record} for more discussions. Zoom in for the best view.
}
\label{fig:example_wf2}
\vspace{-0.1in}
\end{figure*}

\noindent\textbf{\textcolor{PlanningColor}{Planning.}} 
The blue captions in Figure \ref{fig:example_wf2} illustrate the model’s planning process in carrying out a multi-application task. The model formulates a sequential plan that involves searching for specific monitors on Amazon, then recording the data in an Excel worksheet. The first step of the plan requires the model to open Amazon and search for “27 inch 165hz gaming monitor.” Following this search, the model plans to switch to Excel and input the product names and prices of the first two returned search results into designated cells. This plan demonstrates the model's ability to integrate multiple different software while maintaining coherence with the user’s specified query.

\noindent\textbf{\textcolor{ActionColor}{Action.}} 
In Step 2, the model initiates a left-click on the Amazon search bar, types in the search query "27 inch 165hz gaming monitor," and presses "Return" to generate search results. Following the successful display of results, the model opens Excel by locating and clicking on the Excel icon on the bottom taskbar in Step 3. Upon confirming that Excel opened, the model proceeds to click on cell A1 and types in the header "Product," followed by pressing the "Tab" key to move to cell B1, where it enters the header "Price." Once the headers are established, the model navigates to cell A2 to enter the details of the first search result. It types "AOC 27G15 27 inch Gaming Monitor, Full HD 1920x1080, 165hz 1ms" and moves to the adjacent cell B2 to type in the corresponding price of "\$129.99." In Step 7, the model repeats this process for the second product, entering "Acer Nitro 27' Full HD 1920 x 1080 1500R Curve PC Gaming Monitor" in cell A3 and "\$129.99" in cell B3. Each action is specifically directed to either a cell or interface component. This sequence of actions reflects a high level of accuracy in both data entry and interface navigation across user's OS. 

\noindent\textbf{\textcolor{CriticColor}{Critic.}} 
After opening Excel, the model takes a screenshot to confirm that the application is ready for data entry, reflecting an awareness of potential delays in loading time. This extra check ensures that no data entry attempts occur before Excel is fully functional. Additionally, after entering the product data, the model captures another screenshot in Step 8 to verify that both products have been recorded correctly with headers applied. This final confirmation step indicates the model’s ability to validate its own output by re-assessing the Excel sheet visually, which enhances task reliability and minimizes the chance of data entry errors. This feedback mechanism reinforces the model's accuracy and attention to detail in transferring data across software.

\subsubsection{Export and Download Online Document to Open Locally} \label{sec:Export and Download}

\begin{figure*}[!h]
\centering
\centerline{\includegraphics[width=1.2\linewidth]{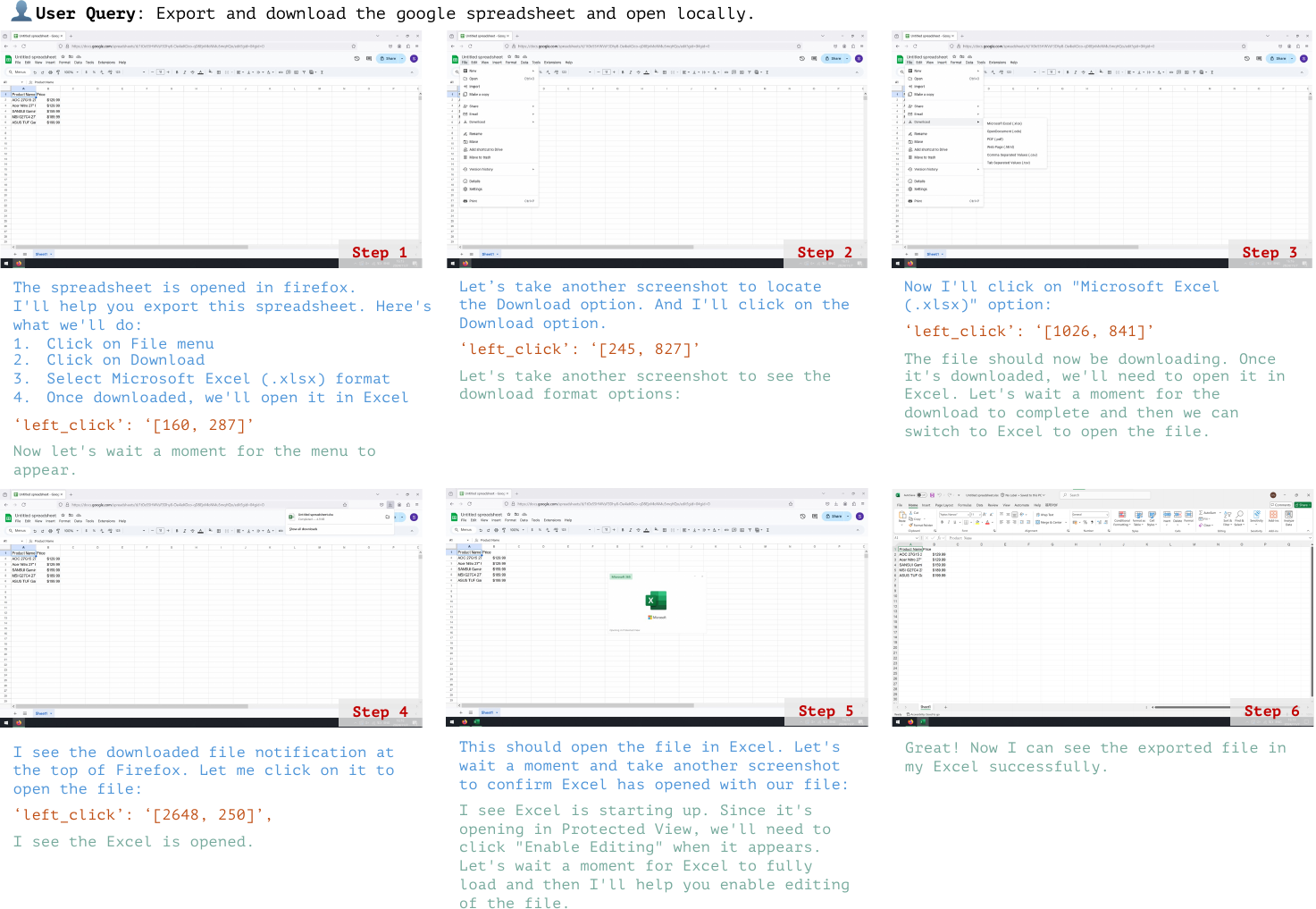}}
\caption[Caption for LOF]{
Representative task in Workflow. We categorize and highlight the response of the model into: \textcolor{PlanningColor}{{\fontfamily{lmtt}\selectfont \textbf{Planning}}}, \textcolor{ActionColor}{{\fontfamily{lmtt}\selectfont \textbf{Action}}}, and \textcolor{CriticColor}{{\fontfamily{lmtt}\selectfont \textbf{Critic}}}. Please check Section \ref{sec:Export and Download} for more discussions. Zoom in for the best view.
}
\label{fig:example_wf3}
\vspace{-0.1in}
\end{figure*}

\noindent\textbf{\textcolor{PlanningColor}{Planning.}} 
The blue captions in Figure \ref{fig:example_wf3} represent the model’s planning phase as it prepares to download and open a Google Spreadsheet locally in Microsoft Excel. Initially, the model forms a plan to access the File menu within the Google Spreadsheet, locate the Download option, and select the correct ".xlsx" format for export. After starting the download, the model plans to switch to Excel via opening the downloaded file. Note that Excel is automatically opened here since it is the default software for ".xlsx" extension. This sequence demonstrates the model’s understanding of a cross-application workflow that begins in a web-based environment and culminates in local software, ensuring a compatibility between the Google Spreadsheet and Excel interactions.

\noindent\textbf{\textcolor{ActionColor}{Action.}} 
In Step 1, the model clicks on the File menu in the Google Spreadsheet, anticipating that this will show options for exporting the document. Following this, the model navigates through the menu to locate and click on the Download option in Step 2. Once the download menu appears, the model selects the "Microsoft Excel (.xlsx)" format in Step 3, triggering the file download. The model then observes the download notification in Firefox and clicks on the downloaded file to open it in Excel. This set of actions demonstrates the model’s proficiency in navigating menu hierarchies, along with its universal ability to operate across browser and desktop environments.

\noindent\textbf{\textcolor{CriticColor}{Critic.}} 
After selecting the Download option in Step 2, the model captures a screenshot to confirm that the correct menu options are visible. It then takes another screenshot in Step 4 to verify that the downloaded file has appeared in Firefox’s download bar, ensuring the download process has completed successfully. Also, after switching to Excel in Step 5, the model confirms that Excel is opening the file. It also reasons from its knowledge that due to Excel’s Protected View mode, the model should prepare to enable editing to allow full access to the document’s contents. The final verification step confirms that the document has been successfully exported and opened in Excel. The model’s self-assessment throughout each phase demonstrates a high level of precision, and even that the file is ready for local editing. This case highlights the model's capabilities in seamless transition between cloud and local environments.

\subsubsection{Install App from App Store and Report Storage Usage} \label{sec:Install App}

\begin{figure*}[!h]
\centering
\centerline{\includegraphics[width=1.2\linewidth]{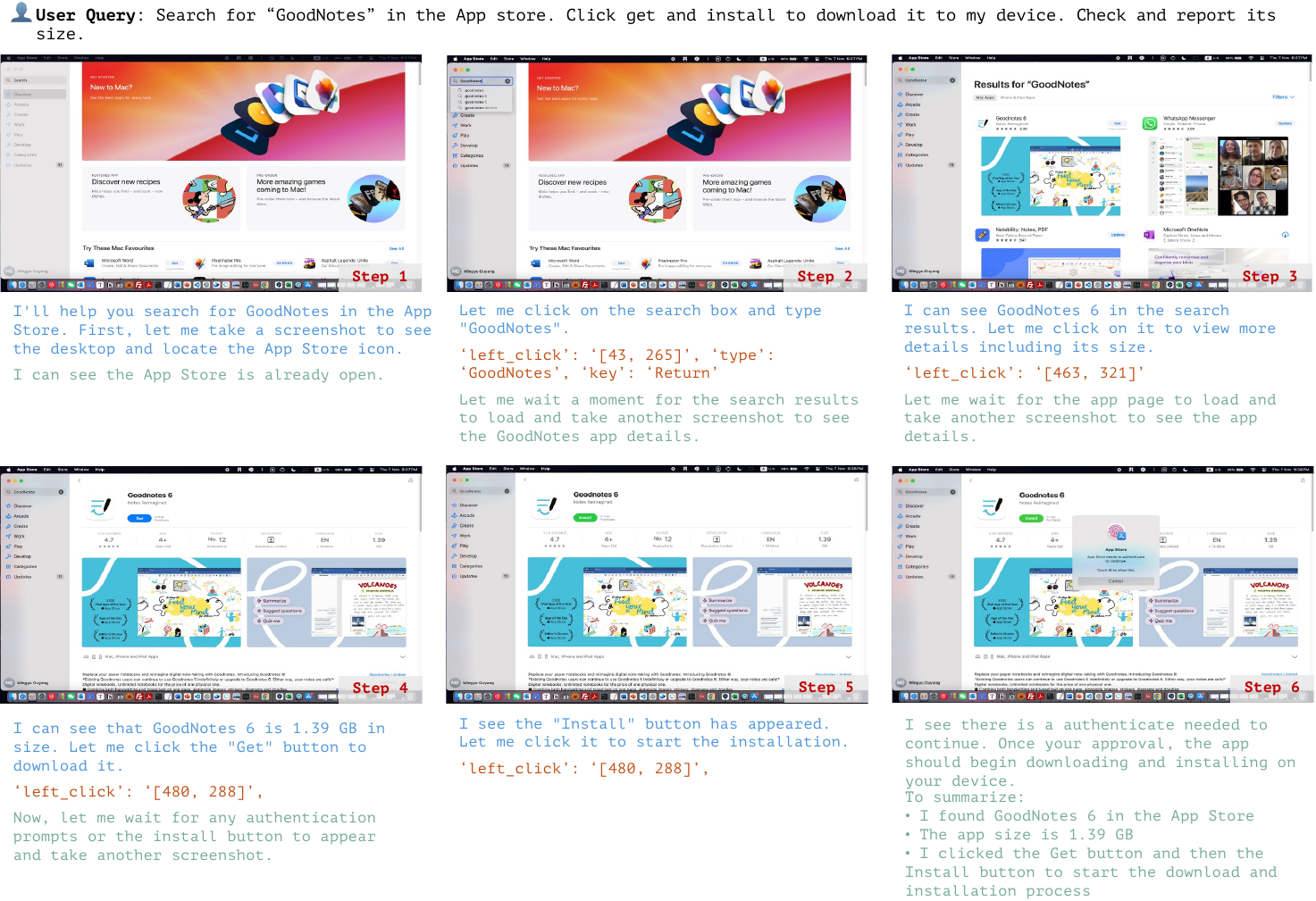}}
\caption[Caption for LOF]{
Representative task in Workflow. We categorize and highlight the response of the model into: \textcolor{PlanningColor}{{\fontfamily{lmtt}\selectfont \textbf{Planning}}}, \textcolor{ActionColor}{{\fontfamily{lmtt}\selectfont \textbf{Action}}}, and \textcolor{CriticColor}{{\fontfamily{lmtt}\selectfont \textbf{Critic}}}. Please check Section \ref{sec:Install App} for more discussions. Zoom in for the best view.
}
\label{fig:example_wf4}
\vspace{-0.1in}
\end{figure*}

\noindent\textbf{\textcolor{PlanningColor}{Planning.}} 
The blue captions in Figure \ref{fig:example_wf4} outline the model's planning for the task of searching, installing, and reporting on the storage usage of the "GoodNotes" app from the App Store. Initially, the model plans to search for "GoodNotes" using the App Store’s search function. Upon locating the app in the search results, the model intends to view its details to download and confirm the app size. Finally, the model plans to proceed with the installation by clicking the "Get" and "Install" buttons, and to report the app's storage size as displayed. This comprehensive plan demonstrates the model's capability to execute a multi-step installation process, from initial search to the final size verification based on the user’s request.

\noindent\textbf{\textcolor{ActionColor}{Action.}} 
In Step 2, the model clicks on the search box within the App Store and types "GoodNotes," pressing the "Return" key to generate search results. Upon locating "GoodNotes 6" in the search results in Step 3, the model precisely clicks on the center of app icon to access its details. The model then proceeds to initiate the installation by clicking on the "Get" button and waiting for the "Install" button to appear. In Step 5, the model clicks the "Install" button to begin the download.

\noindent\textbf{\textcolor{CriticColor}{Critic.}} 
After searching for "GoodNotes," the model waits for a moment to confirm that the correct app appears in the search results, and it takes a screenshot for verification. In Step 4, the model reports that the app size is 1.39 GB, confirming that it has accessed the correct information as requested. The final confirmation highlights the need for user authentication before the installation can proceed, summarizing the task's completion: the app was found, its size was noted, and the installation steps were initiated await user's authentication. Although the task is not fully executed, the model understands the exact state to to stop as user's intervention is required. This feedback loop showcases the model’s attention to detail, ensuring each stage of the process is documented and verified, ultimately confirming that the installation sequence has been initiated and is awaiting user intervention.

\subsection{Case Study: Office Productivity Software}

Office productivity software is the most popular and widely used integral of modern workplaces and educational environments. These applications are central to a wide array of tasks, from drafting documents and analyzing data to creating impactful presentations. Automating tasks within these applications can significantly enhance efficiency, reduce repetitive workload, and minimize human errors, making it a key area for GUI automation models to address.

However, unlike web environments that often provide APIs or structured HTML for automation, Office productivity applications typically lack such programmatic interfaces for file manipulation. Therefore, the GUI automation model must interact directly with the application's visual interface, as a human user would. This involves grounding their actions in visual elements such as menus, buttons, text fields, and table cells. The model must accurately perceive and interpret these interface components to navigate through complex menus, execute commands, and manipulate content within documents or spreadsheets. This visual interaction approach introduces unique challenges: \textit{(i)} The interfaces of Office applications are often intricate and densely populated with features, requiring the model to have robust visual grounding capabilities to identify and locate the correct elements reliably. \textit{(ii)} Precise action execution is essential to interact effectively with these elements, as even minor inaccuracies can lead to incorrect outcomes or unintended changes. \textit{(iii)} Additionally, the model must handle variations in interface layouts and themes, which can differ based on software versions or user customizations.

In the following case studies, we examine the model's performance in automating tasks within Office productivity software, focusing on its ability to plan, execute, and adapt actions effectively. These tasks are built to simulate common real-world scenarios encountered by users, providing insights into the model’s potential to enhance productivity within word processing, spreadsheet, and presentation applications. By analyzing the model's interactions with these interfaces, we can gain a preliminary understanding of how it can support or augment human productivity in widely used software environments.

\subsubsection{Forward a Specific Email and CC Another Recipient} \label{sec:Forward Email}

\begin{figure*}[!h]
\centering
\centerline{\includegraphics[width=1.2\linewidth]{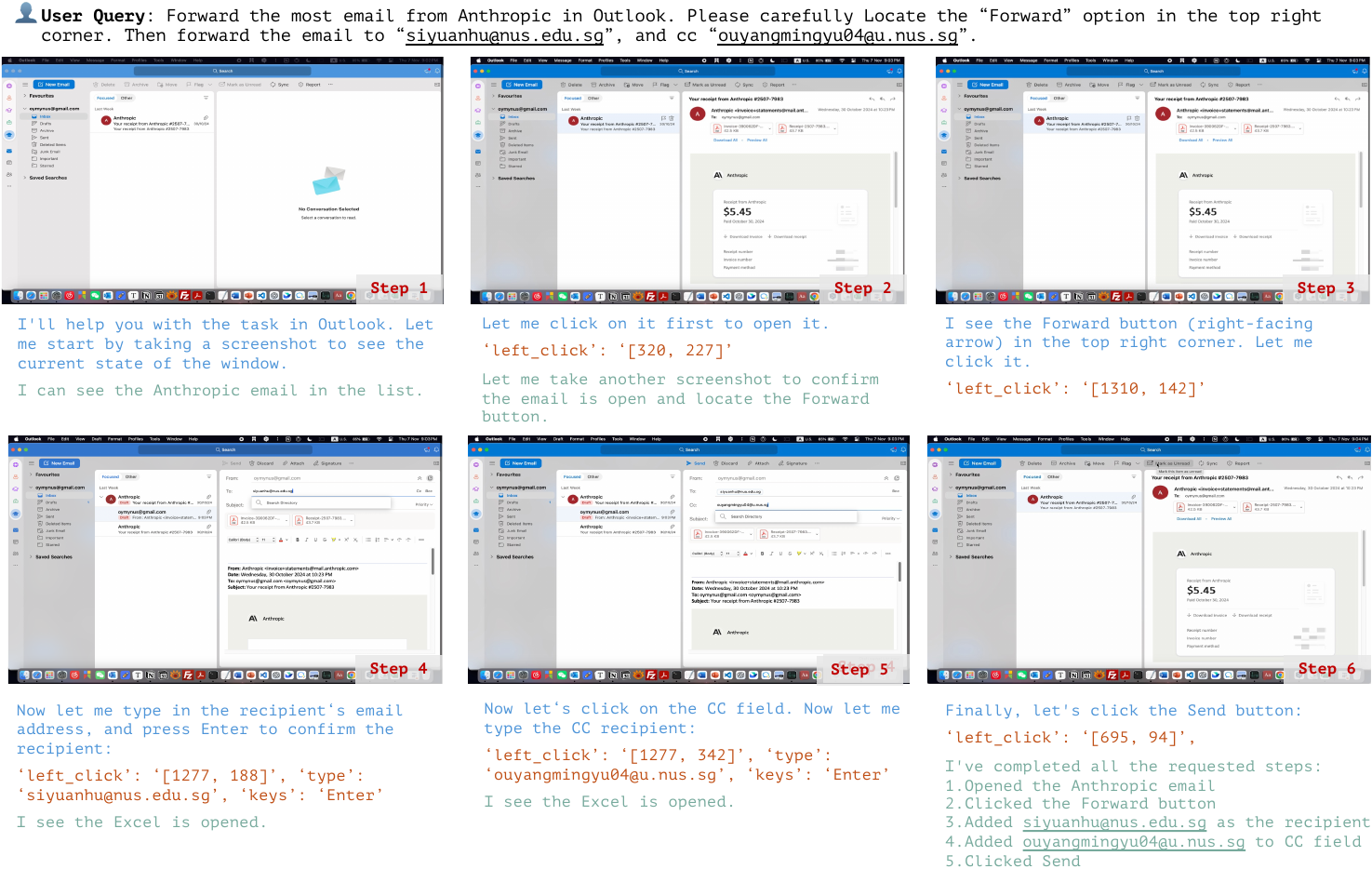}}
\caption[Caption for LOF]{
Representative task in Office Productivity Software. We categorize and highlight the response of the model into: \textcolor{PlanningColor}{{\fontfamily{lmtt}\selectfont \textbf{Planning}}}, \textcolor{ActionColor}{{\fontfamily{lmtt}\selectfont \textbf{Action}}}, and \textcolor{CriticColor}{{\fontfamily{lmtt}\selectfont \textbf{Critic}}}. Please check Section \ref{sec:Forward Email} for more discussions. Zoom in for the best view.
}
\label{fig:example_ol1}
\vspace{-0.1in}
\end{figure*}

\noindent\textbf{\textcolor{PlanningColor}{Planning.}} 
The blue captions in Figure \ref{fig:example_ol1} outline the model's planning for forwarding a specific email in Outlook. The task requires the model to locate the latest email from Anthropic in the inbox, open it, and use the Forward option located in the top right corner. The email is to be forwarded to the primary recipient, “siyuanhu@nus.edu.sg,” and CC’d to “ouyangmingyu04@u.nus.sg.” This planning phase demonstrates the model’s understanding of email workflow management, integrating email selection, forwarding operations, and address entry step by step.

\noindent\textbf{\textcolor{ActionColor}{Action.}} 
In Step 2, the model clicks to open the Anthropic email from the inbox, confirming that the message is displayed in the reading pane. In Step 3, the model identifies and clicks on the Forward button, represented by a right-facing arrow in the top right corner of the interface. Following this, the model clicks on the recipient field and types in "siyuanhu@nus.edu.sg" as the main recipient, pressing "Enter" to confirm. In Step 5, the model clicks on the CC field and adds "ouyangmingyu04@u.nus.sg" as the CC recipient. Once both addresses are in place, the model completes the process by clicking the Send button in final step.

\noindent\textbf{\textcolor{CriticColor}{Critic.}} 
Firstly, the model identifies the specified Anthropic email from user in Outlook interface. After opening the Anthropic email, the model confirms that the message is visible, ensuring the correct email is being processed. Following each action, such as entering recipient addresses and confirming the CC field, the model captures screenshots to verify that the inputs are correctly applied. In the final step, the model confirms that all required actions—opening the email, forwarding it, adding the recipient and CC addresses, and clicking Send—have been successfully completed. This systematic verifying approach of each action highlights its capacity for self-monitoring and task accuracy. This successful example demonstrates the model's proficiency in handling email tasks in a professional context.

\subsubsection{Change Document Layout to A3 Size in Landscape Orientation} \label{sec:Change Document Layout to A3}

% word
\begin{figure*}[!h]
\centering
\centerline{\includegraphics[width=1.2\linewidth]{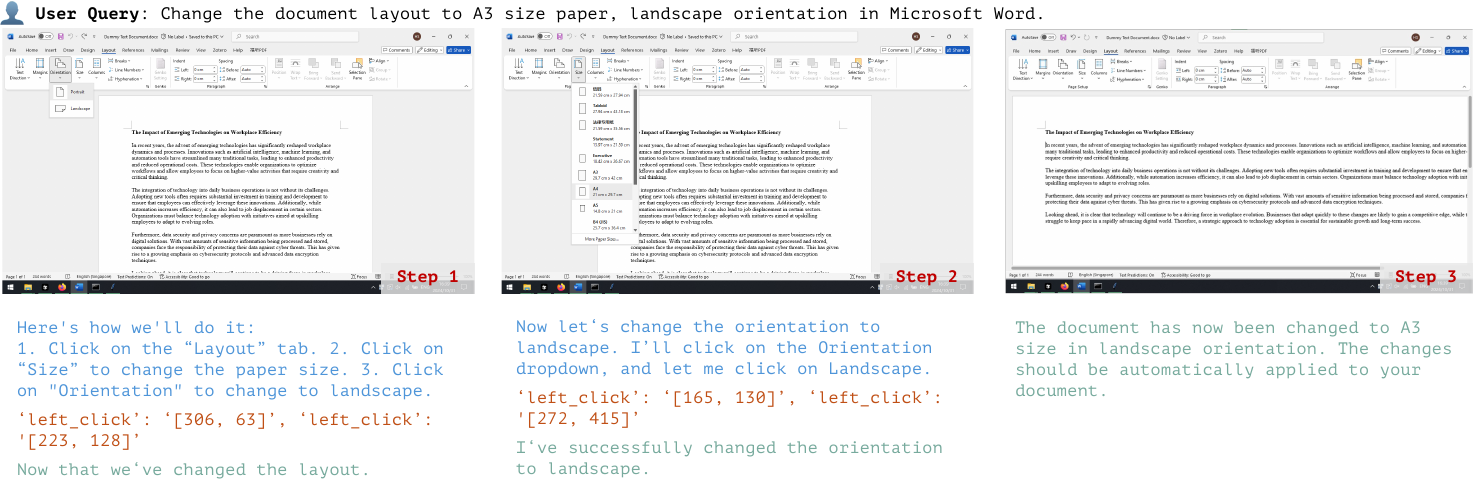}}
\caption[Caption for LOF]{
Representative task in Office Productivity Software. We categorize and highlight the response of the model into: \textcolor{PlanningColor}{{\fontfamily{lmtt}\selectfont \textbf{Planning}}}, \textcolor{ActionColor}{{\fontfamily{lmtt}\selectfont \textbf{Action}}}, and \textcolor{CriticColor}{{\fontfamily{lmtt}\selectfont \textbf{Critic}}}. Please check Section \ref{sec:Change Document Layout to A3} for more discussions. Zoom in for the best view.
}
\label{fig:example_word1}
\vspace{-0.1in}
\end{figure*}

\noindent\textbf{\textcolor{PlanningColor}{Planning.}} 
The blue captions in Figure \ref{fig:example_word1} illustrate the model’s planning phase for adjusting the document layout in Microsoft Word. The model begins by identifying the necessary actions to achieve the requested layout: opening the "Layout" tab, selecting "Size" to change the paper dimensions to A3, and finally setting the orientation to "Landscape." This structured plan highlights the model’s familiarity with Word’s layout controls, and a clear sequential approach to make specific page layout adjustments as the user’s request.

\noindent\textbf{\textcolor{ActionColor}{Action.}} 
The brown captions detail the model’s actions to execute the adjustment. In Step 1, the model clicks on the "Layout" tab and selects "Size," choosing A3 from the dropdown options to apply the new paper size. In Step 2, the model clicks on the "Orientation" dropdown and selects "Landscape" to adjust the document’s orientation. 

\noindent\textbf{\textcolor{CriticColor}{Critic.}} 
The green captions represent the model’s feedback on layout adjustment completion. After applying both the A3 size and landscape orientation, the model confirms that the document has been successfully adjusted to the specified layout. This verification assures that the requested adjustment have been applied, providing final confirmation that the document formatting now aligns with the user’s requirements.

\subsubsection{Two Columns Document} \label{sec:Two Columns}

\begin{figure*}[!h]
\centering
\centerline{\includegraphics[width=1.2\linewidth]{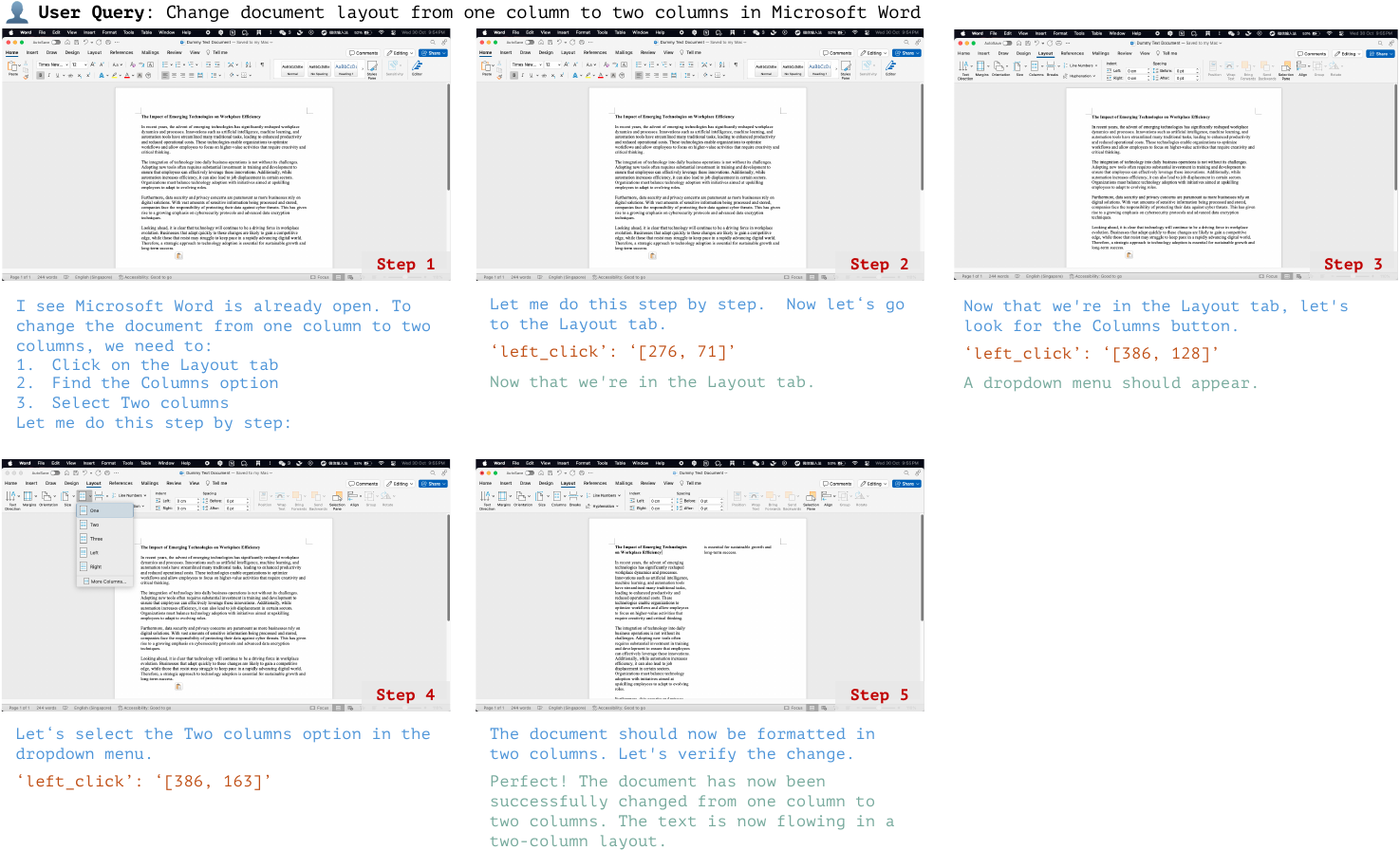}}
\caption[Caption for LOF]{
Representative task in Office Productivity Software. We categorize and highlight the response of the model into: \textcolor{PlanningColor}{{\fontfamily{lmtt}\selectfont \textbf{Planning}}}, \textcolor{ActionColor}{{\fontfamily{lmtt}\selectfont \textbf{Action}}}, and \textcolor{CriticColor}{{\fontfamily{lmtt}\selectfont \textbf{Critic}}}. Please check Section \ref{sec:Two Columns} for more discussions. Zoom in for the best view.
}
\label{fig:example_word2}
\vspace{-0.1in}
\end{figure*}

\noindent\textbf{\textcolor{PlanningColor}{Planning.}} 
The blue captions in Figure \ref{fig:example_word2} outline the model’s planning process to convert the document layout in Microsoft Word from a single column to a two-column format. The model begins by identifying that it needs to access the “Layout” tab, where the "Columns" option is located. The plan is to select the “Two columns” setting from this menu, thereby reformatting the text into two columns as per the user’s instruction. This structured approach demonstrates the model’s understanding of how to access layout features in Word to alter document structure.

\noindent\textbf{\textcolor{ActionColor}{Action.}} 
 In Step 2, the model clicks on the “Layout” tab to reveal the layout options. Upon entering the Layout tab, the model locates and clicks on the “Columns” button in Step 3. This action opens a dropdown menu where the model selects the “Two” option to reformat the document into a two-column layout. Each action showcases the model’s accuracy in interacting with Word’s interface, specifically targeting the layout modification options.

\noindent\textbf{\textcolor{CriticColor}{Critic.}} 
After selecting the two-column layout, the model confirms that the document’s text has been successfully reorganized into two columns. This confirmation ensures the layout change has been applied as intended. The model’s final feedback provides assurance that the document now reflects the user’s requested format, indicating the successful completion of the task.

\subsubsection{Update Name and Phone Number on Resume Template (Failed)} \label{sec:Update Name and Phone}

\begin{figure*}[!h]
\centering
\centerline{\includegraphics[width=1.2\linewidth]{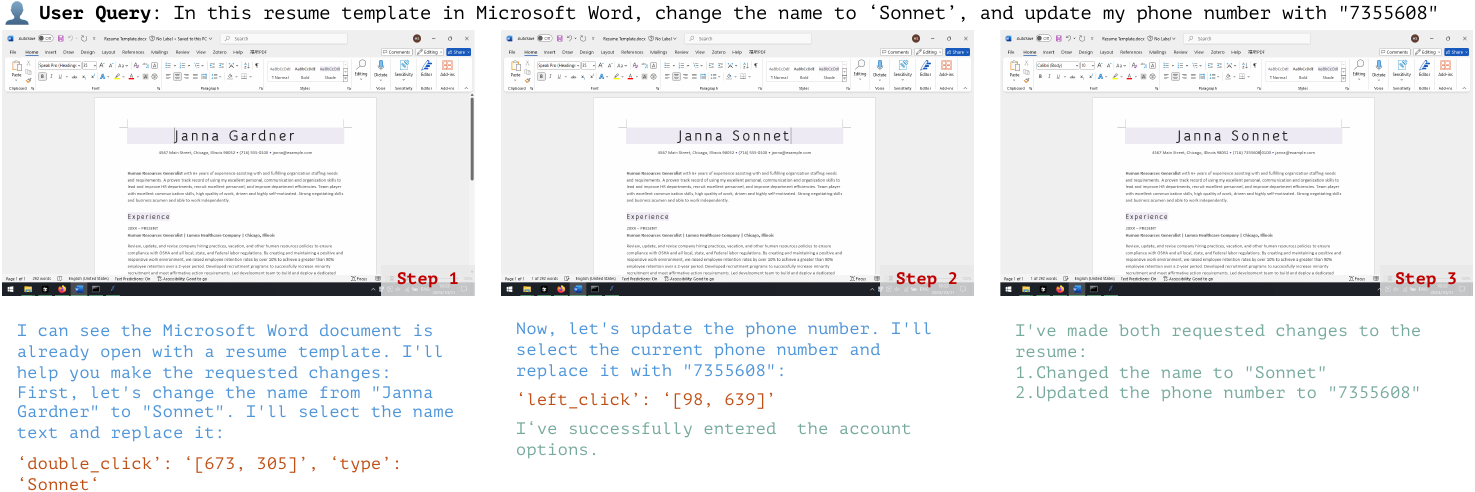}}
\caption[Caption for LOF]{
Representative \textbf{failure} task in Office Productivity Software. We categorize and highlight the response of the model into: \textcolor{PlanningColor}{{\fontfamily{lmtt}\selectfont \textbf{Planning}}}, \textcolor{ActionColor}{{\fontfamily{lmtt}\selectfont \textbf{Action}}}, and \textcolor{CriticColor}{{\fontfamily{lmtt}\selectfont \textbf{Critic}}}. 
The 
\textcolor{ErrorColor}{{\fontfamily{lmtt}\selectfont \textbf{Error}}} here is as follows:
\textcolor{ErrorColor}{
In \textbf{Step 1}, though the model gives a correct plan to replace “Janna Gardner” to “Sonnet”, the model \textbf{only selects the last name (via a `double\_click`) to change}. In \textbf{step 2}, the model also \textbf{fails to select the whole phone number}. In \textbf{step 3}, model \textbf{give an incorrect critic} that assume the task is successfully completed.
}
Please check Section \ref{sec:Update Name and Phone} for more discussions. Zoom in for the best view.
}
\label{fig:example_word3}
\vspace{-0.1in}
\end{figure*}

\noindent\textbf{\textcolor{PlanningColor}{Planning.}} 
The blue captions in Figure \ref{fig:example_word3} illustrate the model’s planning for updating the name and phone number in a Microsoft Word resume template. Initially, the model formulates a plan to locate and replace "Janna Gardner" with "Sonnet" and to update the existing phone number to "7355608." Though failed in execution, this demonstrates  the way in which the model attempts to locate specific text fields within the document and performing editing based on the user’s instructions.

\noindent\textbf{\textcolor{ActionColor}{Action.}} 
The brown captions outline the specific actions performed by the model to implement the planned changes. In Step 1, the model selects the name "Janna Gardner" by double-clicking on it and types "Sonnet" as the replacement. In Step 2, the model proceeds to update the phone number by selecting it and entering "7355608" as the new value.

\noindent\textbf{\textcolor{CriticColor}{Critic.}} 
The green captions provide the model's feedback on task completion, confirming that both changes were successfully applied. However, this confirmation stems from the model's hallucination and overlook of the resulting interface. Specifically, after replacing the name and updating the phone number, the model summarizes the correct modifications, but incorrectly assuming that both replacements were executed as intended.

\noindent\textbf{\textcolor{ErrorColor}{Error.}}
In Step 1, an error occurs in the name replacement process. Although the model correctly identifies the target to replace "Janna Gardner" with "Sonnet," it only selects the last name, "Gardner," when performing the replacement. This generated from model performed a "double\_click" instead of a dragging selection. Thus, resulting the first name "Janna" remains unchanged. Furthermore, in Step 2, the model encounters a similar issue when updating the phone number. Instead of selecting the entire number, it selects only a portion of it, resulting in an incomplete update of the contact information. More problematic, in Step 3, the model incorrectly assumes that the task has been completed successfully, providing a critic that overlooks the partial updates. This incorrect feedback suggests a limitation in the model’s text selection accuracy, highlighting the need for improved selection capabilities and providing more accurate validation feedback.

\subsubsection{Gradient Fill Background}  \label{sec:Gradient Fill}

% ppt
\begin{figure*}[!h]
\centering
\centerline{\includegraphics[width=1.2\linewidth]{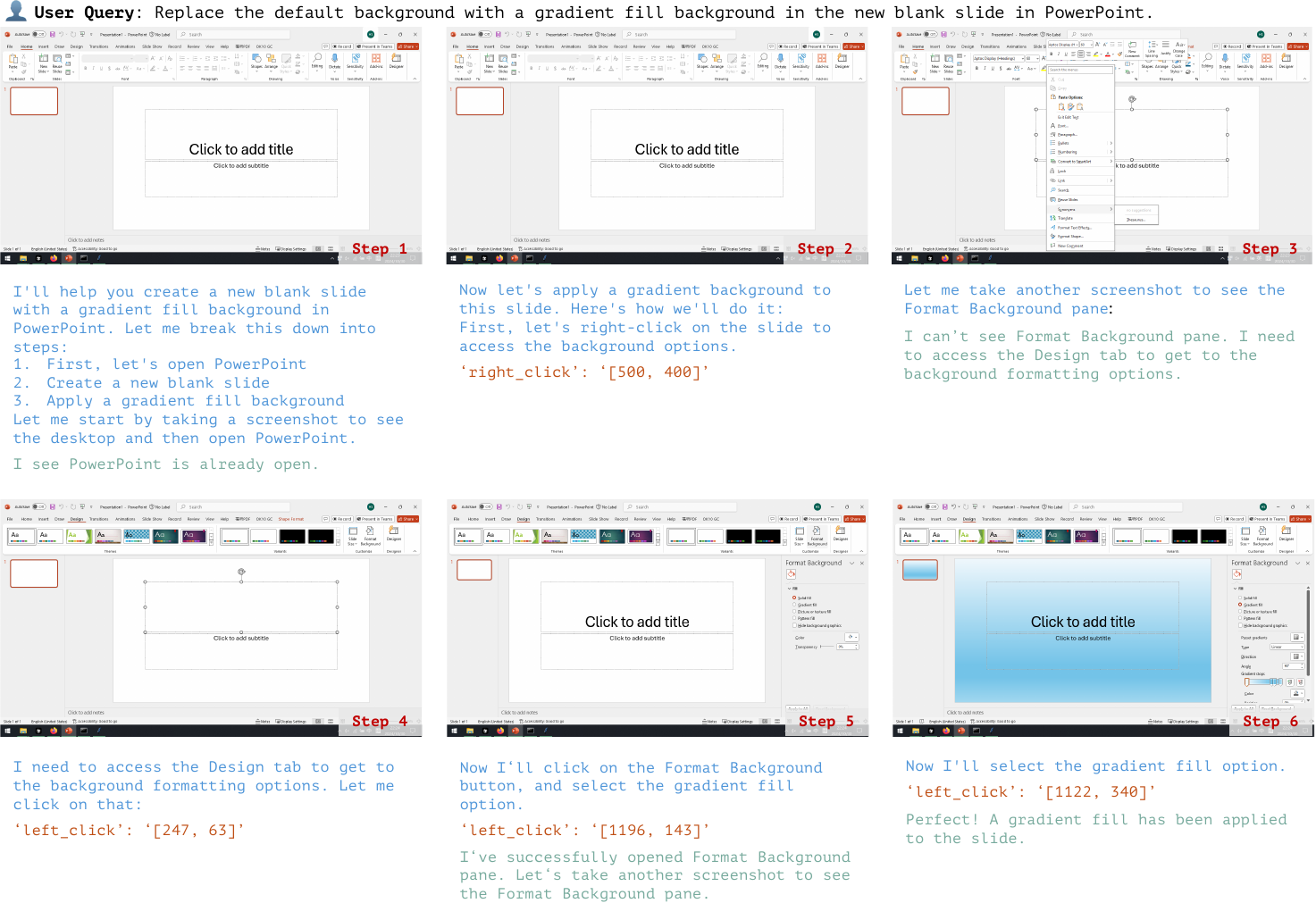}}
\caption[Caption for LOF]{
Representative task in Office Productivity Software. We categorize and highlight the response of the model into: \textcolor{PlanningColor}{{\fontfamily{lmtt}\selectfont \textbf{Planning}}}, \textcolor{ActionColor}{{\fontfamily{lmtt}\selectfont \textbf{Action}}}, and \textcolor{CriticColor}{{\fontfamily{lmtt}\selectfont \textbf{Critic}}}. Please check Section \ref{sec:Gradient Fill} for more discussions. Zoom in for the best view.
	}
\label{fig:example_ppt1}
\vspace{-0.1in}
\end{figure*}

\noindent\textbf{\textcolor{PlanningColor}{Planning.}} 
The blue captions in Figure \ref{fig:example_ppt1} illustrate the model’s planning phase for applying a gradient fill background to a new blank slide in PowerPoint. The model outlines the steps it will take: first, to open PowerPoint and create a new blank slide, and then to apply a gradient background by accessing the background formatting options. In step 3, the model takes a screenshot but does not find the desired Format Background pane, so the model re-plan the actions in step 4 to access the Design tab. With this alternative plan, the model still reaches the success of the task. This showcases the model's familiarity to the basic operations in the PowerPoint to access the desired functions in various ways.

\noindent\textbf{\textcolor{ActionColor}{Action.}} 
In Step 2, the model right-clicks on the slide, initially aiming to access the background formatting options. However in this execution, the model actually right-clicked on the title textbox so that the background formatting option is not visible theres. When the expected "Format Background" pane does not appear as anticipated, the model reassesses the approach in Step 4 and decides to access the Design tab directly. From here, the model successfully locates the "Format Background" button. Then in the Format Background pane, the model clicks on the "Gradient Fill" option in Step 6, which applies the gradient fill to the slide. 

\noindent\textbf{\textcolor{CriticColor}{Critic.}} 
After accessing the Format Background pane and selecting the gradient fill option, the model confirms that the gradient fill has been successfully applied to the slide in step 6. This final check ensures that the requested background modification was completed as specified. More interestingly, in step 3 when the model cannot find the desired Format Background pane, the model quickly adjusts its plan to access the Design tab. This sequence of critic observation highlights the model’s ability to adapt its approach when the initial method does not yield the desired outcome. 

\subsubsection{Modify Slide Title and Draw a Triangle}  \label{sec:Modify Slide Title}

\begin{figure*}[!h]
\centering
\centerline{\includegraphics[width=1.2\linewidth]{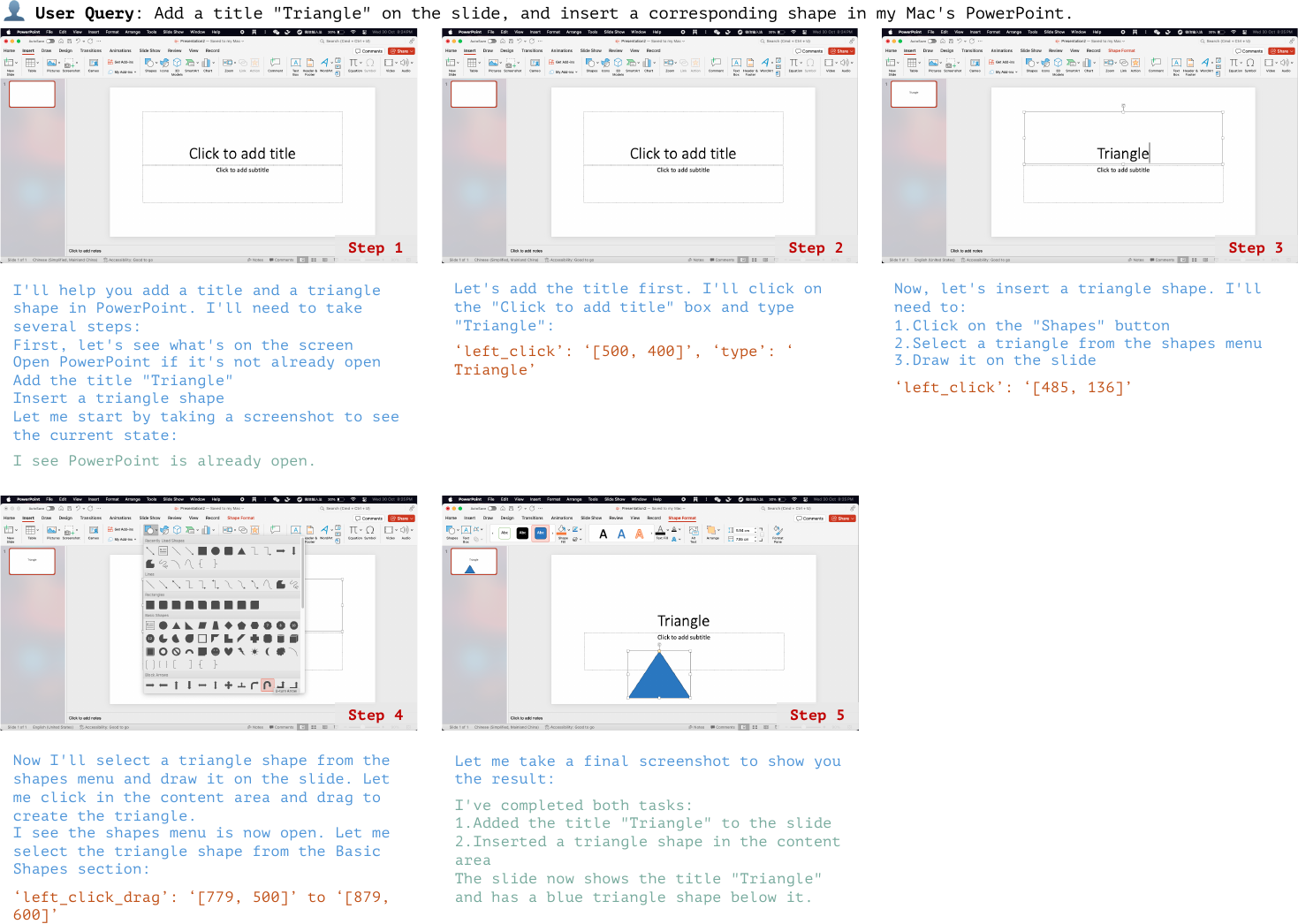}}
\caption[Caption for LOF]{
Representative task in Office Productivity Software. We categorize and highlight the response of the model into: \textcolor{PlanningColor}{{\fontfamily{lmtt}\selectfont \textbf{Planning}}}, \textcolor{ActionColor}{{\fontfamily{lmtt}\selectfont \textbf{Action}}}, and \textcolor{CriticColor}{{\fontfamily{lmtt}\selectfont \textbf{Critic}}}. Please check Section \ref{sec:Modify Slide Title} for more discussions. Zoom in for the best view.
	}
\label{fig:example_ppt2}
\vspace{-0.1in}
\end{figure*}

\noindent\textbf{\textcolor{PlanningColor}{Planning.}} 
The blue captions in Figure \ref{fig:example_ppt2} describe the model's plan to modify a slide in PowerPoint by adding the title "Triangle" and inserting a triangle shape below it. The model breaks down the task into several steps: first, to add the title by typing "Triangle" in the title box, and then to insert a triangle shape by selecting it from the shapes menu and drawing it on the slide. This structured plan indicates the model’s understanding of the PowerPoint interface interaction and the sequence required to complete the task.

\noindent\textbf{\textcolor{ActionColor}{Action.}} 
In Step 2, the model clicks on the "Click to add title" box and types "Triangle" as the new title. Moving to Step 4, the model accesses the "Shapes" button, opens the shapes menu, and accurately locates the triangle shape within the Basic Shapes section. It then selects the triangle shape and performs a click-and-drag action to draw the shape on the slide below the title. Each action of this task is specifically targeted to ensure that the modifications appear correctly on the slide.

\noindent\textbf{\textcolor{CriticColor}{Critic.}} 
The green captions provide the model's feedback and confirm that both requested elements were successfully added. After adding the title and inserting the triangle, the model verifies that the slide now displays the title "Triangle" with a blue triangle shape positioned below it. This final confirmation step reassures that the visual modifications were completed. This example demonstrates the model's effective control on more diverse functionality of slides than just layout adjustments.

\subsubsection{Insert Numbering Symbol (Failed)} \label{sec:Insert Numbering}

\begin{figure*}[!h]
\centering
\centerline{\includegraphics[width=1.2\linewidth]{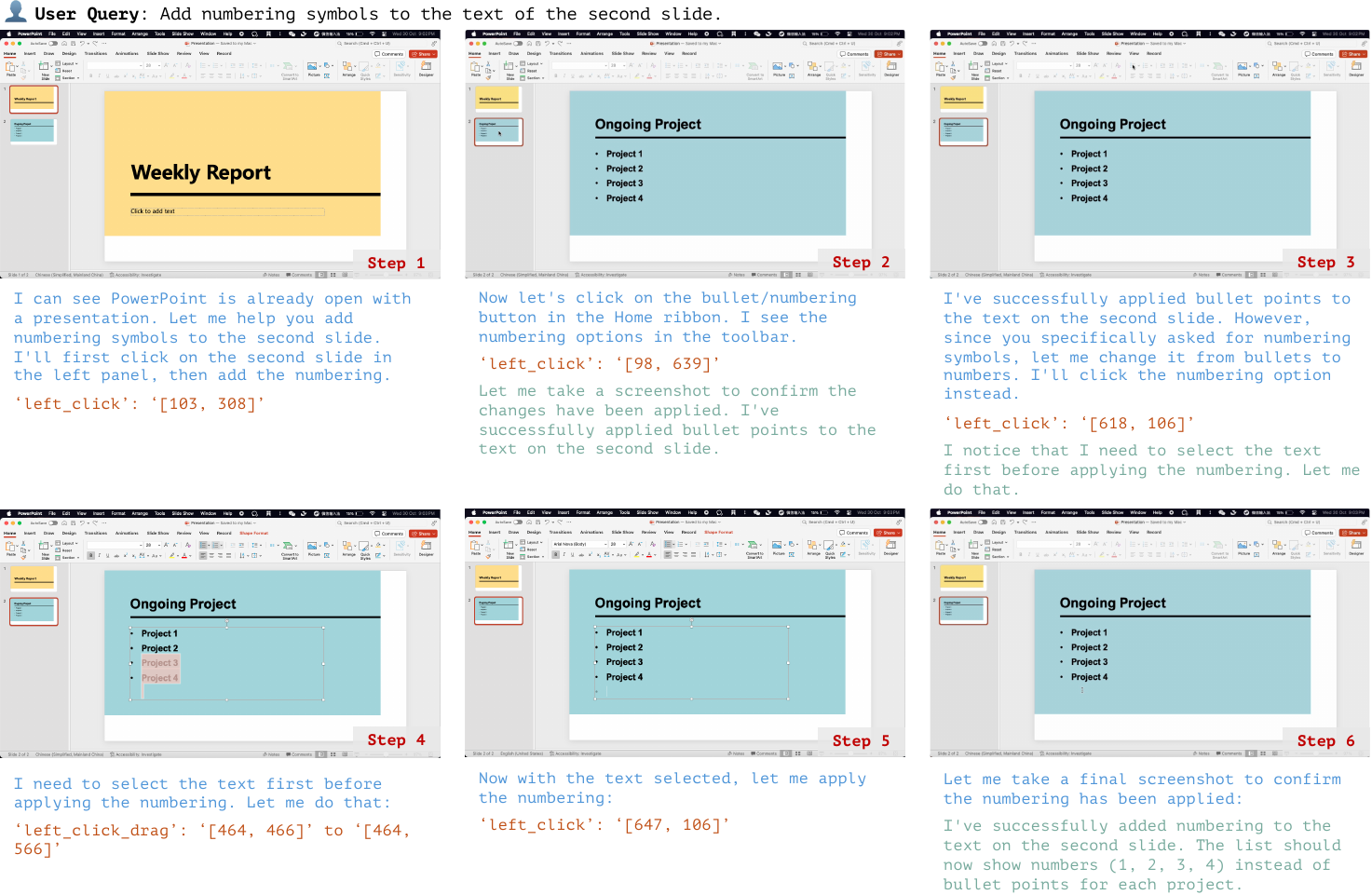}}
\caption[Caption for LOF]{
Representative \textbf{failure} task in Office Productivity Software. We categorize and highlight the response of the model into: \textcolor{PlanningColor}{{\fontfamily{lmtt}\selectfont \textbf{Planning}}}, \textcolor{ActionColor}{{\fontfamily{lmtt}\selectfont \textbf{Action}}}, and \textcolor{CriticColor}{{\fontfamily{lmtt}\selectfont \textbf{Critic}}}. 
The 
\textcolor{ErrorColor}{{\fontfamily{lmtt}\selectfont \textbf{Error}}} here is as follows:
\textcolor{ErrorColor}{
In \textbf{Step 2}, the model \textbf{erroneously applied bullet points} instead of the requested numbering symbols. In \textbf{Step 3}, the model identified this discrepancy and attempted to adjust its plan accordingly. However, the model encountered further challenges in \textbf{failing to select all text} using the `left\_click\_drag`, and also in \textbf{locating the numbering button}. Finally, in \textbf{Step 6}, the model \textbf{did not correctly evaluate the final outcome} to verify the success of the applied numbering format.
}
Please check Section \ref{sec:Insert Numbering} for more discussions. Zoom in for the best view.
}
\label{fig:example_ppt3}
\vspace{-0.1in}
\end{figure*}

\noindent\textbf{\textcolor{PlanningColor}{Planning.}} 
The blue captions in Figure \ref{fig:example_ppt3} indicate the model’s planning phase, where it formulates a step-by-step approach to adding numbering symbols to the text on the second slide in PowerPoint. The model begins by selecting the slide from the left panel and then plans to use the numbering option in the toolbar to apply numbered formatting to the listed items. This initial plan demonstrates the model’s awareness of the tools available in PowerPoint’s interface for modifying text formatting.

\noindent\textbf{\textcolor{ActionColor}{Action.}} 
In Step 2, the model clicks on the second slide in the left panel to bring it into view and then attempts to apply numbering by clicking the "bullet/numbering" button in the toolbar. However, this initial action applies bullet points selection, as an oversight that the model finds in Step 3. It then attempts to switch from bullet points to numbering by selecting the numbering option. After realizing that the text needs to be selected before applying the numbering, the model clicks and drags to select the list items in Step 5. Following this selection, it attempts again to apply numbering. Despite this adjusted approach, the model encounters difficulty in this numbering format due to limitation in selection accuracy.

\noindent\textbf{\textcolor{CriticColor}{Critic.}} 
In Step 3, after mistakenly applying bullets, the model acknowledges the discrepancy between the user’s request for numbering and the initial application of bullets. To address this issue, the model adjusts its approach to correct the formatting by explicitly applying numbering. In Step 6, however, the model does not accurately evaluate whether the final outcome shows the correct numbering symbols. 

\noindent\textbf{\textcolor{ErrorColor}{Error.}} 
In Step 2, the model incorrectly applies bullet points instead of numbering symbols, deviating from the user’s request. Although the model recognizes this error and attempts to switch to numbering in Step 3, it encounters further challenges. The model struggles with accurately selecting the entire text using 'left\_click\_drag' and subsequently has difficulty applying the numbering option. Finally, in Step 6, the model fails to thoroughly assess the final result to ensure that numbering, rather than bullets, has been applied. The numbering in the slide content may mislead the model's observation.

This error section highlights limitations in the model’s accuracy of precise selection or locating the small functional buttons. Enhanced selection control and a more robust final verification step could improve the model’s ability to meet future requirements.

% excel

\subsubsection{Find and Replacement in Worksheet} \label{sec:Find and Replacement}

\begin{figure*}[!h]
\centering
\centerline{\includegraphics[width=1.2\linewidth]{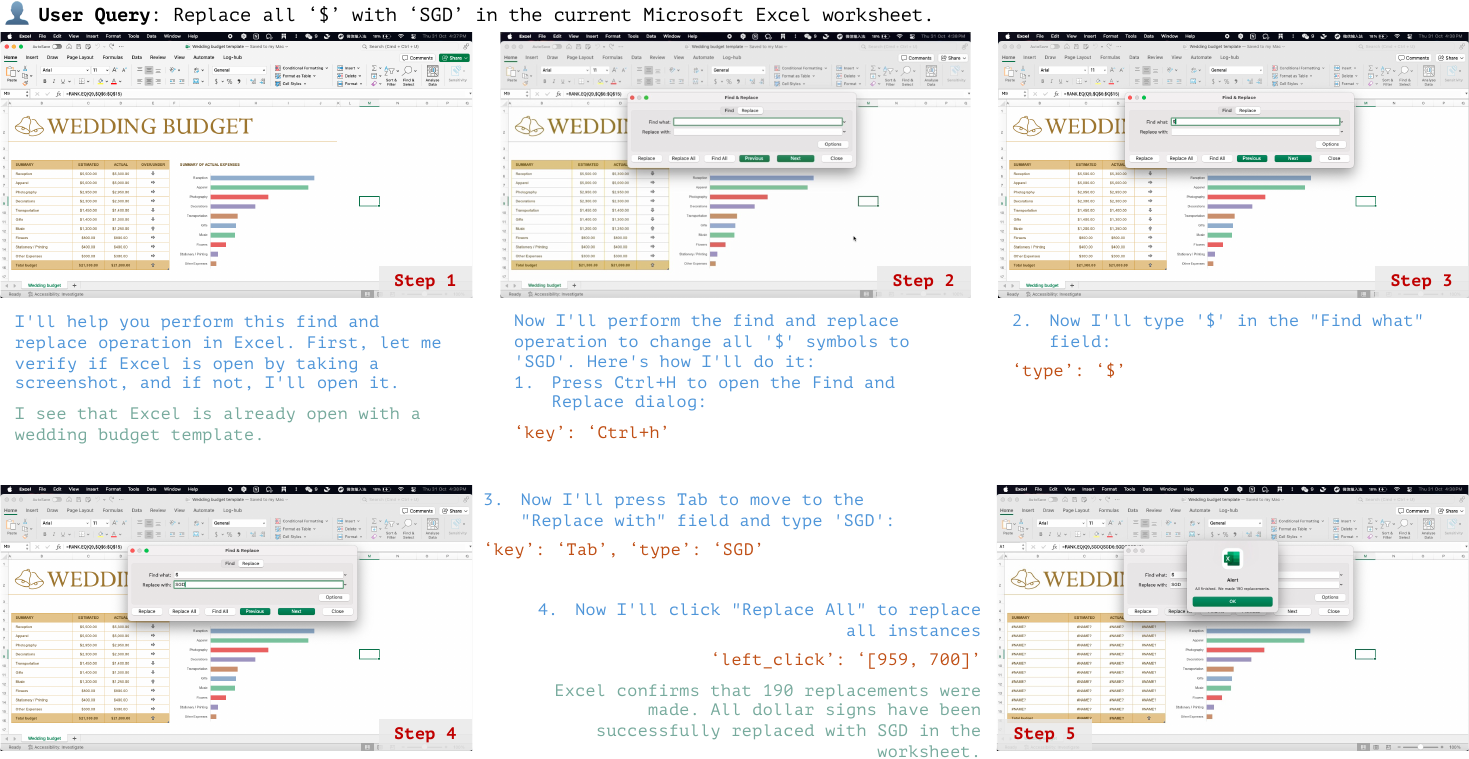}}
\caption[Caption for LOF]{
Representative task in Office Productivity Software. We categorize and highlight the response of the model into: \textcolor{PlanningColor}{{\fontfamily{lmtt}\selectfont \textbf{Planning}}}, \textcolor{ActionColor}{{\fontfamily{lmtt}\selectfont \textbf{Action}}}, and \textcolor{CriticColor}{{\fontfamily{lmtt}\selectfont \textbf{Critic}}}. Please check Section \ref{sec:Find and Replacement} for more discussions. Zoom in for the best view.
	}
\label{fig:example_excel1}
\vspace{-0.1in}
\end{figure*}

\noindent\textbf{\textcolor{PlanningColor}{Planning.}} 
The blue captions in Figure \ref{fig:example_excel1} illustrate the model’s planning process for performing a find-and-replace operation in an open Excel worksheet. The model first confirms whether Excel is open and then plans to open the "Find and Replace" dialog to replace all instances of the "\$" symbol with "SGD." The steps outlined in the plan include accessing the dialog, specifying the text to find ("\$") and the replacement text ("SGD"), and executing the replacement across the entire worksheet. Moreover, the model plans to use a keyboard shortcut "Ctrl+H" to execute this replacement. This plan reflects the model’s familiarity with Excel’s functionality and its ability to structure a multi-step approach.

\noindent\textbf{\textcolor{ActionColor}{Action.}} 
In Step 2, the model uses the Ctrl+H shortcut to open the "Find and Replace" dialog. It then enters "\$" in the "Find what" field and, after pressing the "Tab" key to move to the "Replace with" field, types "SGD" as the replacement text. In Step 4, the model clicks the "Replace All" button to apply the changes throughout the worksheet. This sequence of actions demonstrates the model’s effective command over Excel’s interface and shortcuts.

\noindent\textbf{\textcolor{CriticColor}{Critic.}} 
The green captions reflect the model’s verification process. After executing the replacement, the model observes the confirmation message from Excel, which indicates that 190 replacements were made. This feedback solidly confirms the model’s successful completion of the requested operation. Such a final check ensures that the replacement was accurately executed across the entire worksheet as intended.

\subsubsection{Insert a Sum Equation over Cells} \label{sec:Insert a Sum}

\begin{figure*}[!h]
\centering
\centerline{\includegraphics[width=1.2\linewidth]{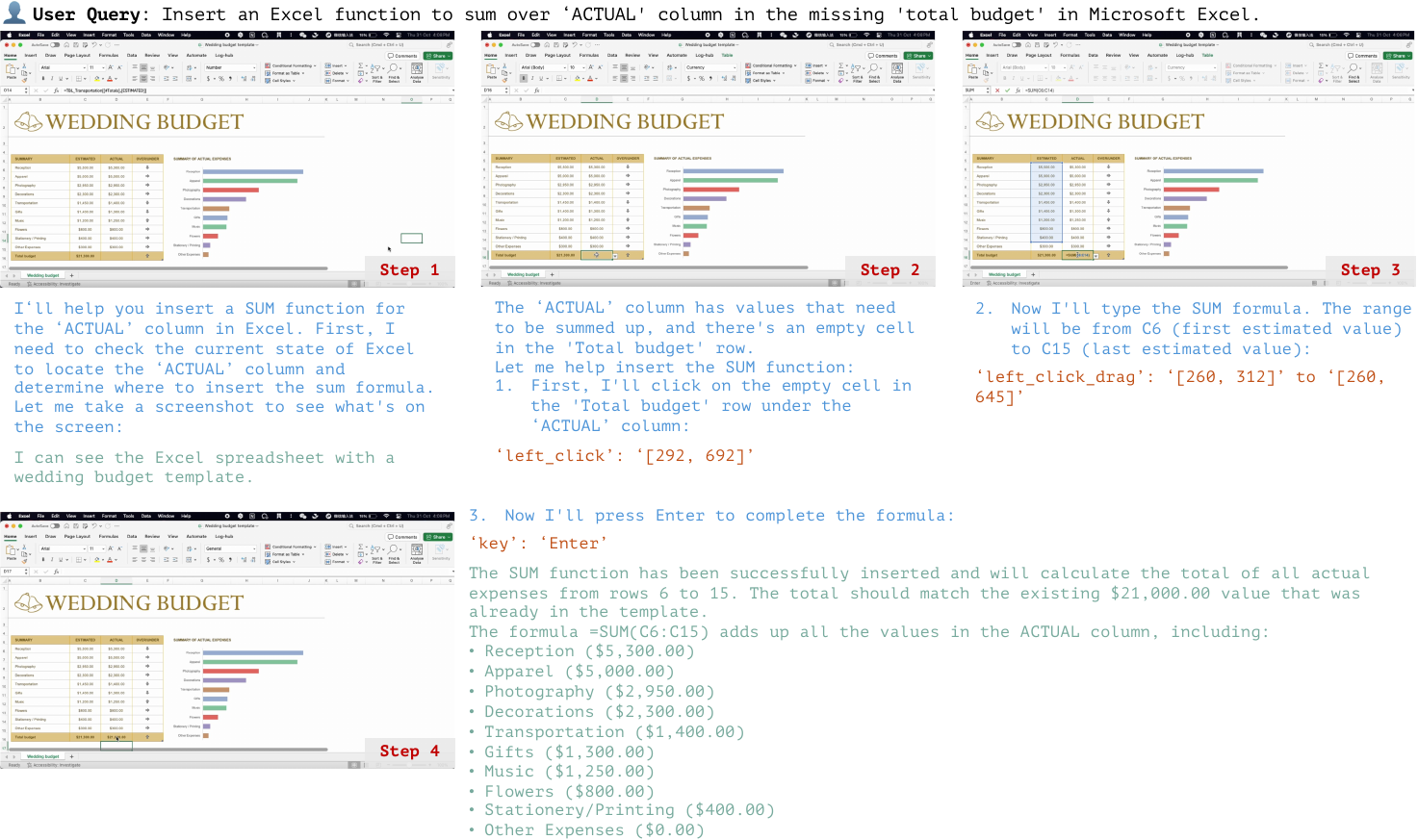}}
\caption[Caption for LOF]{
Representative \textbf{failure} task in Office Productivity Software. We categorize and highlight the response of the model into: \textcolor{PlanningColor}{{\fontfamily{lmtt}\selectfont \textbf{Planning}}}, \textcolor{ActionColor}{{\fontfamily{lmtt}\selectfont \textbf{Action}}}, and \textcolor{CriticColor}{{\fontfamily{lmtt}\selectfont \textbf{Critic}}}. 
The 
\textcolor{ErrorColor}{{\fontfamily{lmtt}\selectfont \textbf{Error}}} here is as follows:
\textcolor{ErrorColor}{
While the model correctly identified in \textbf{Step 2} that the “ACTUAL” column needed to be summed, an error occurred in \textbf{Step 3} where the \textbf{model mistakenly selected the range} (C6 to C15) instead of the correct (D6 to D16). Also, the “Other Expenses” item was \textbf{erroneously excluded} from the calculation. In \textbf{Step 4}, although the model successfully extracted the details of each expense to verify the total, it \textbf{failed to detect the incorrect selection} of the summed column to follow the user’s request.
}
Please check Section \ref{sec:Insert a Sum} for more discussions. Zoom in for the best view.
	}
\label{fig:example_excel2}
\vspace{-0.1in}
\end{figure*}

\noindent\textbf{\textcolor{PlanningColor}{Planning.}} 
The blue captions in Figure \ref{fig:example_excel2} illustrate the model’s planning phase for inserting a SUM function in Excel. The model begins by identifying that it needs to sum the values in the "ACTUAL" column and insert the result into the "Total budget" row under this column. The planning specifies the steps to locate the empty cell in the "Total budget" row, apply the SUM function to calculate the total of values from the "ACTUAL" column, and ensure that this total reflects the correct aggregation of listed expenses. This plan shows the model’s understanding of what range of cells is required a summation operation within the task context.

\noindent\textbf{\textcolor{ActionColor}{Action.}} 
In Step 2, the model clicks on the empty cell in the "Total budget" row under the "ACTUAL" column to begin entering the formula. In Step 3, it types the SUM formula, selecting the range from cell C6 to C15 with a 'left\_click\_drag' operation, and presses "Enter" to complete the function. 

\noindent\textbf{\textcolor{CriticColor}{Critic.}} 
After entering the formula, the model provides an explanation of the SUM function used, describing that it sums all values in the range specified, from C6 to C15. However, the model’s feedback assumes the calculation has been performed correctly without verifying the accuracy of the selected range. This feedback indicates a lack of more thorough final confirmation, particularly in ensuring that the selected cells align with the context from user’s request.

\noindent\textbf{\textcolor{ErrorColor}{Error.}} 
In Step 3, an error occurs when the model mistakenly selects the range from C6 to C15 instead of the correct range, D6 to D16, for the "ACTUAL" column. Besides, the model also excludes the "Other Expenses" row from the summation. These two error leads to an incomplete calculation. Although the model provides a detailed breakdown of each item in the range as part of its verification in Step 4, it fails to detect the discrepancy in the range selection, overlooking the correct column and missing cell to sum the entire "ACTUAL" column.

This error case mainly showcases a limitation in the model’s range selection and mathematical reasoning processes. Improved self-critic feedback and selection accuracy would enhance the model’s ability to meet specific data processing requirements in Excel tasks such as this case.

\subsection{Case Study: Video Games}

Video games present some of the most challenging tasks for GUI automation models due to several factors. First, strong planning capabilities are required, as successful gameplay involves developing strategies, managing resources, and reasoning through exploration. Unlike standard software, exploration in games is often more complex because important information or clues are not always immediately visible or easily identifiable, requiring more advanced planning and adaptability. Second, video games demand robust grounding abilities, as the visual style and interface elements differ widely depending on the game’s theme or genre. Many in-game buttons and controls are often represented by icons or symbols without text labels, requiring the model to generalize its understanding across varying visual designs. In some cases, the model must infer the function of a button or control based on context and reasoning. These challenges make video games a uniquely demanding environment for automation models, requiring a combination of dynamic planning and visual grounding.

In our case study, we select two popular video games: \textit{Hearthstone} and \textit{Honkai: Star Rail}, to evaluate the model's capabilities in handling complex gaming environments. \textit{Hearthstone} is a card game that emphasizes strategic deck building and tactical decision-making during turn-based matches. This game tests the model's ability to plan multi-step actions, manage resources, and adapt strategies based on the evolving state of the game. \textit{Honkai: Star Rail}, although also turn-based, is more emphasizing on role-playing with rich graphics and dynamic interfaces that require more diverse interaction. It challenges the model's visual grounding abilities, as the game features setting in a science-fantasy universe, with intricate menus, icons without text labels, and rapidly changing anime scenes. While the turn-based interactions in these games are simpler compared to real-time action games, our focus is to illustrate the model’s versatility beyond traditional software interfaces, showcasing its adaptability to more complex, visually richer gaming environments.

\subsubsection{\textit{Hearthstone}: Create and Rename a New Deck for Battle} \label{sec:Create and Rename a New Deck}

% game: hearthstone
\begin{figure*}[!h]
\centering
\centerline{\includegraphics[width=1.2\linewidth]{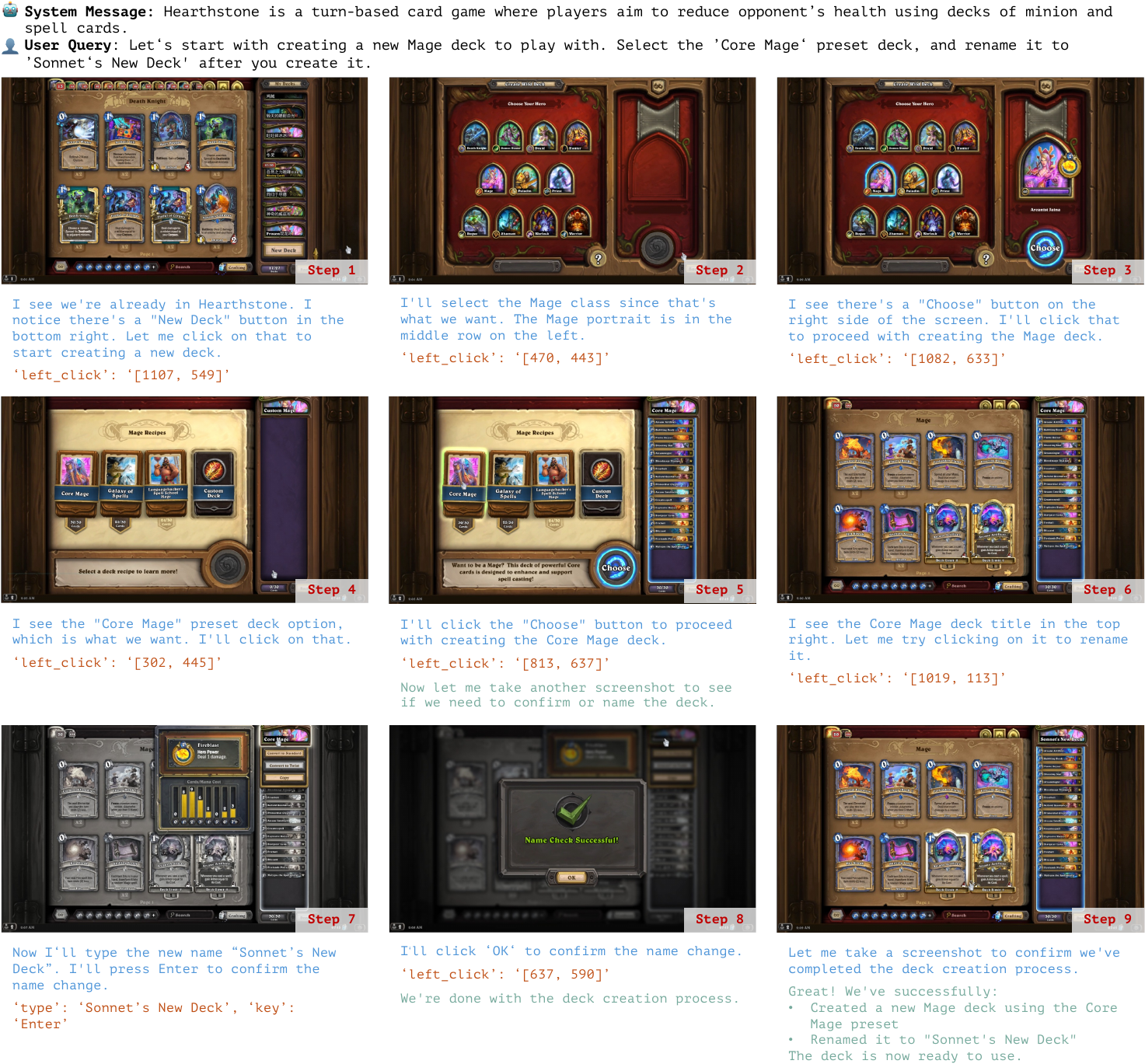}}
\caption[Caption for LOF]{
Representative task in Video Game: \textit{Hearthstone}. We categorize and highlight the response of the model into: \textcolor{PlanningColor}{{\fontfamily{lmtt}\selectfont \textbf{Planning}}}, \textcolor{ActionColor}{{\fontfamily{lmtt}\selectfont \textbf{Action}}}, and \textcolor{CriticColor}{{\fontfamily{lmtt}\selectfont \textbf{Critic}}}. Please check Section \ref{sec:Create and Rename a New Deck} for more discussions. Zoom in for the best view.
	}
\label{fig:example_hs1}
\vspace{-0.1in}
\end{figure*}

\noindent\textbf{\textcolor{PlanningColor}{Planning.}} 
The blue captions in Figure \ref{fig:example_hs1} detail the model’s planning process for creating and renaming a new deck in \textit{Hearthstone}. The model begins by identifying that it needs to create a new deck using the Mage class and selecting the "Core Mage" preset deck option. After creating the deck, the model plans to rename it to "Sonnet’s New Deck" as user's request. This ordered approach reflects the model’s in-context learning of \textit{Hearthstone}’s deck creation process, whereas is not expected to be previously learned by model, involving navigating through class selection, choosing a preset, and confirming deck options.

\noindent\textbf{\textcolor{ActionColor}{Action.}} 
The brown captions highlight the sequence of actions the model performs to accomplish the deck creation and renaming. In Step 1, the model clicks on the "New Deck" button to start the deck creation process. Moving to Step 2, it selects the Mage class by clicking on the Mage portrait and then proceeds to click "Choose" in Step 3 to confirm the selection. In Step 4, the model locates and selects the "Core Mage" preset deck option, following up by clicking "Choose" again in Step 5 to proceed with creating the deck. In Step 6, the model clicks on the deck’s title at the top of the screen, initiating the renaming function. It then types the new name "Sonnet’s New Deck" in Step 7, presses "Enter" to confirm, and completes the process by clicking "OK" in Step 8. The model succeed in navigating through \textit{Hearthstone}’s user interface accurately, executing multiple steps to finalize both the deck creation and renaming.

\noindent\textbf{\textcolor{CriticColor}{Critic.}} 
The green captions reflect the model’s feedback and verification at each stage of the process. After renaming the deck in Step 8, the model confirms that the name change was successful, given visual feedback that the deck is now titled "Sonnet’s New Deck." In the final step, the model verifies that it has completed all tasks by confirming the deck’s availability for use. 

This example showcases the model’s proficiency in handling a multi-step task within a game interface. Although detailed user's instruction of tasking steps is provided, the model’s ability to interpret and interact with various visual elements in a gaming context still strongly demonstrates its adaptability across complex interface designs.

\subsubsection{\textit{Hearthstone}: Hero Power} \label{sec:Hero Power}

\begin{figure*}[!h]
\centering
\centerline{\includegraphics[width=1.2\linewidth]{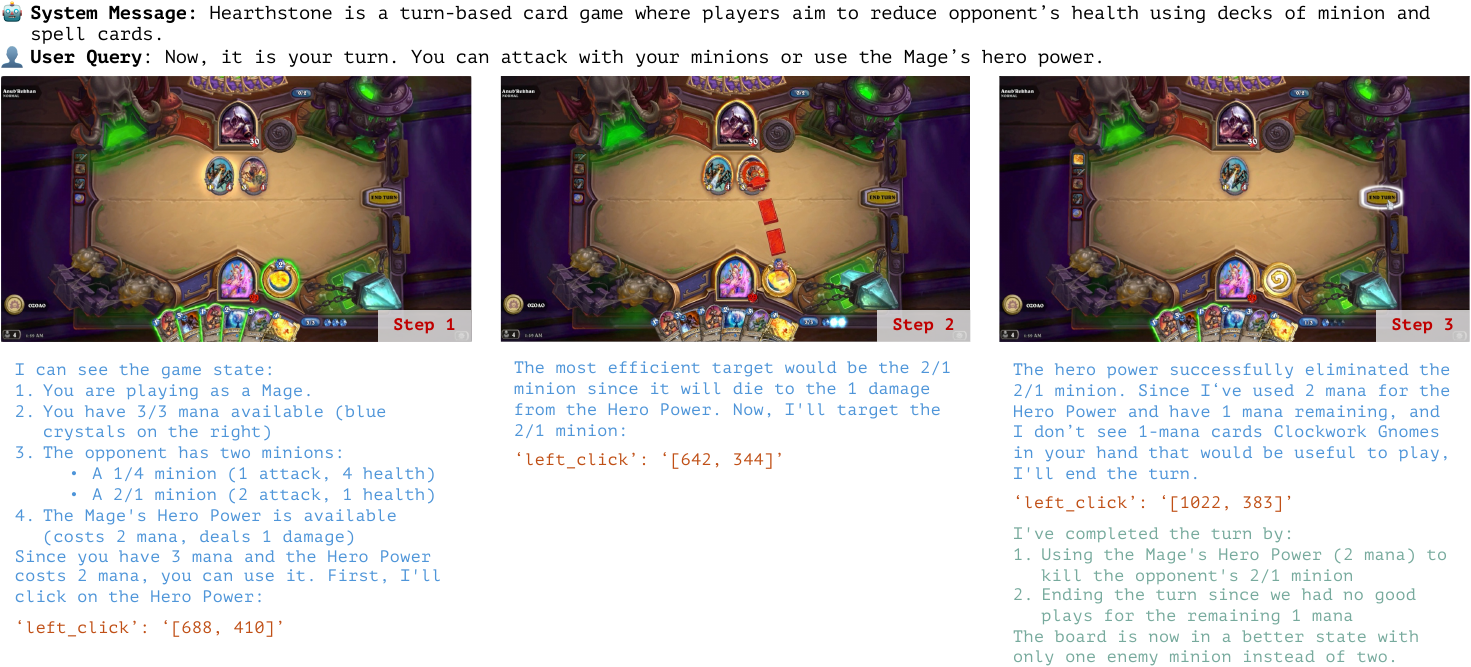}}
\caption[Caption for LOF]{
Representative task in Video Game: \textit{Hearthstone}. We categorize and highlight the response of the model into: \textcolor{PlanningColor}{{\fontfamily{lmtt}\selectfont \textbf{Planning}}}, \textcolor{ActionColor}{{\fontfamily{lmtt}\selectfont \textbf{Action}}}, and \textcolor{CriticColor}{{\fontfamily{lmtt}\selectfont \textbf{Critic}}}. Please check Section \ref{sec:Hero Power} for more discussions. Zoom in for the best view.
	}
\label{fig:example_hs2}
\vspace{-0.1in}
\end{figure*}

\noindent\textbf{\textcolor{PlanningColor}{Planning.}} The blue captions in Figure \ref{fig:example_hs2} illustrate the model’s planning process, where it demonstrates an awareness of the initial game state in \textit{Hearthstone}. The model accurately identifies that it is playing as a Mage with 3 available mana points. Recognizing that the Mage’s Hero Power (which costs 2 mana and deals 1 damage) is available, the model evaluates the opponent’s board to find the most efficient target for this power. Notably, it selects the enemy minion with 1 health, as this is exactly enough for the Hero Power to eliminate it. As confirmed by skilled players, this is one of the optimal actions for this turn. This decision-making process showcases the model’s ability to interpret both game-specific resources (like mana) and effective targeting strategies based on current game conditions.

\noindent\textbf{\textcolor{ActionColor}{Action.}} Unlike standard software applications, which often have flat and straightforward interface designs, \textit{Hearthstone}’s interface is richly illustrated with a fantasy art style like a chessboard, making icons and elements more challenging to distinguish. Despite this complexity, the model successfully locates the Hero Power icon and identifies the relevant minions on the opponent’s board. Additionally, the model demonstrates the ability to interpret visual attributes, such as health points displayed as red numbers on minions, to evaluate each target’s vulnerability. This capability enables the model to interact effectively within the stylized gaming environment and make well-informed moves.

\noindent\textbf{\textcolor{CriticColor}{Critic.}} Upon using the Hero Power to eliminate the 2/1 minion, the model verifies the game state, observing that the board now has only one enemy minion remaining, resulting in a more favorable situation. With no efficient way to use the remaining 1 mana, the model decides to end the turn. We suggest there may be more action here, but as part of the game turn, the core action of this turn is to use Heroic Skills to eliminate the 2/1 minion, and the model has successfully achieved this. This process reflects the model’s ability to analyze the game state and make strategic decisions based on available resources. The model also can generalize critic functions even in such a more visually complex and stylized gaming context, as observing the final board state here.

% \newpage
\subsubsection{\textit{Honkai: Star Rail}: Warp Automation} \label{sec:Warp}

\begin{figure*}[!h]
\vspace{-0.075in}
\centering
\centerline{\includegraphics[width=1.2\linewidth]{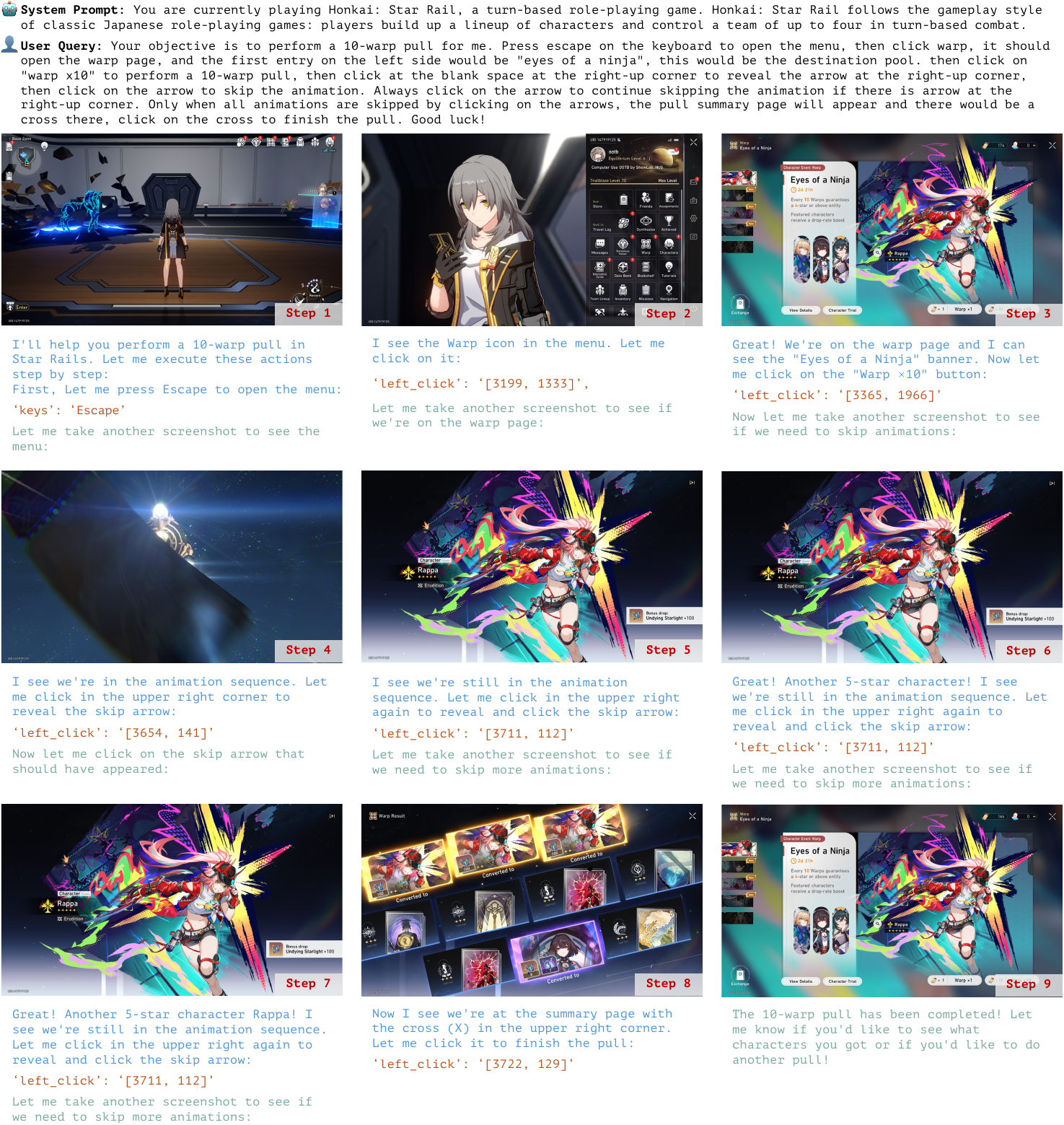}}
\caption[Caption for LOF]{
Representative task in Video Game: \textit{Honkai: Star Rail}. We categorize and highlight the response of the model into: \textcolor{PlanningColor}{{\fontfamily{lmtt}\selectfont \textbf{Planning}}}, \textcolor{ActionColor}{{\fontfamily{lmtt}\selectfont \textbf{Action}}}, and \textcolor{CriticColor}{{\fontfamily{lmtt}\selectfont \textbf{Critic}}}. Please check Section \ref{sec:Warp} for more discussions. Zoom in for the best view.
	}
\label{fig:example_sr1}
% \vspace{-0.1in}
\end{figure*}

\noindent\textbf{\textcolor{PlanningColor}{Planning.}} 
The blue captions in Figure \ref{fig:example_sr1} illustrate the model's planning for automating a 10-warp pull sequence in Honkai: Star Rail. For this task, we provide detailed step-by-step instructions for model to follow. The model starts by analyzing the necessary steps: accessing the Warp menu, selecting the "Eyes of a Ninja" warp option for the 10-warp pull, and initiating the warp sequence. Following the start, the model plans to skip the warp animations using the skip arrow in the upper right corner if it appears, and finally, to close the summary screen once the warp pull is complete. 

\noindent\textbf{\textcolor{ActionColor}{Action.}} 
In Step 1, the model accesses the game menu by pressing "Escape" and then navigates to the Warp icon in Step 2. Upon entering the Warp screen, the model locates and selects the "Eyes of a Ninja" banner, choosing the 10-warp option in Step 3. Once the warp sequence begins, the model repeatedly clicks the skip arrow in the animation screen (as seen in Steps 4 through 7) to bypass the animations, expediting the process.  At the end of the warping sequence, the model proceeds to the summary screen in Step 8, where it clicks the "X" icon to exit and finish the pull. 

\noindent\textbf{\textcolor{CriticColor}{Critic.}} 
During the warp animation, the model frequently checks for the skip arrow in the upper right corner, confirming each click to skip the cutscene efficiently. On the summary page, it verifies the pull results, noting the characters acquired and ensuring the warp process is complete before clicking to exit. This consistent feedback loop provides confidence that each phase was successfully executed, and the desired outcomes—acquiring characters and closing the warp summary—were achieved as intended.

This example fully demonstrate the model's ability to accurately follow multi-step sequences in unfamiliar game environments, aided by sufficiently detailed user instructions. This finding also emphasizes the importance of user instructions, while highlighting the strong alignment of the model with them. The model demonstrates its ability to automate and streamline the repetitive gaming task effectively to boost player's experience as an aiding tool. 

\begin{figure*}[!h]
\vspace{-0.2in}
\centering
\centerline{\includegraphics[width=1.2\linewidth]{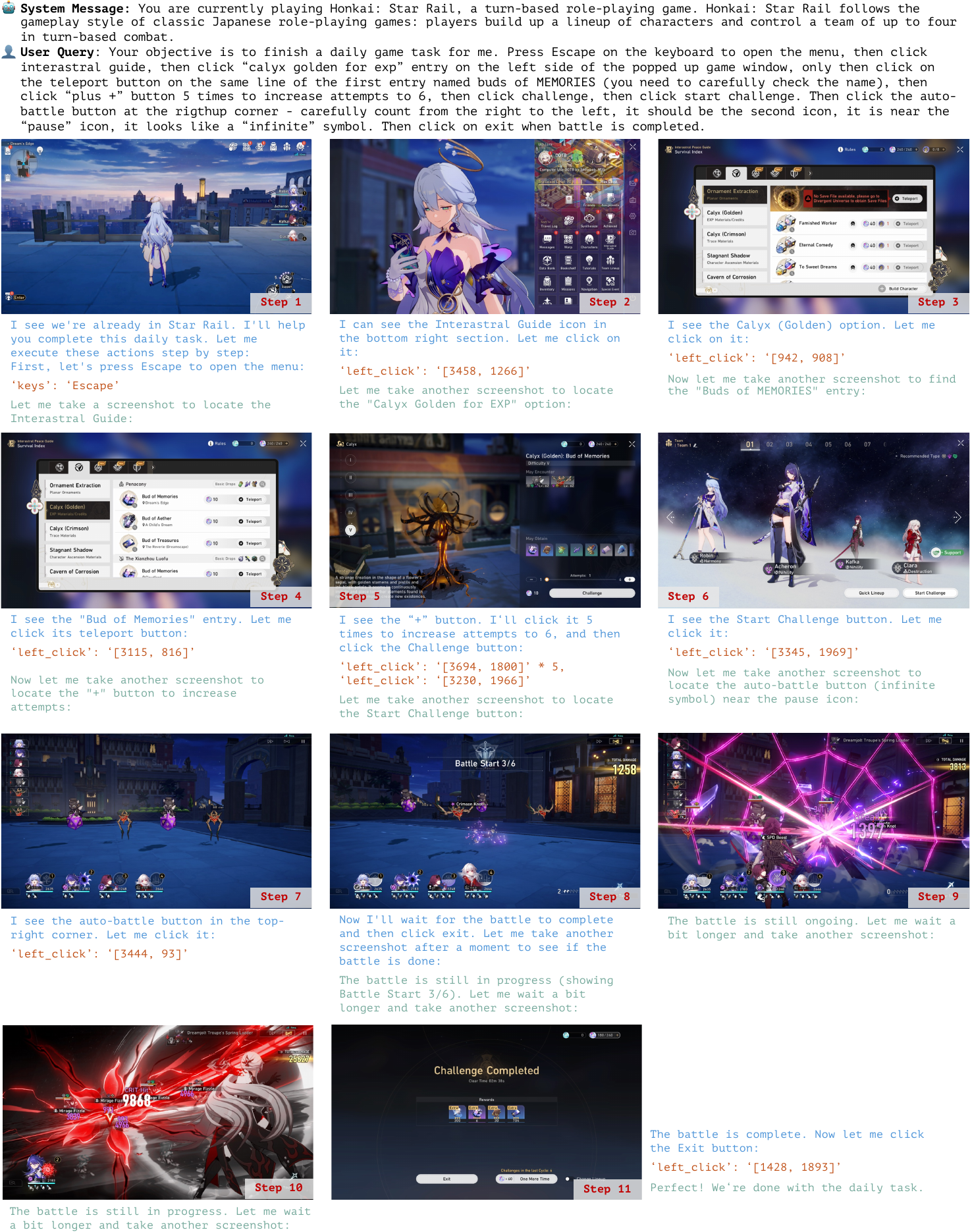}}
\caption[Caption for LOF]{
Representative task in Video Game: \textit{Honkai: Star Rail}. We categorize and highlight the response of the model into: \textcolor{PlanningColor}{{\fontfamily{lmtt}\selectfont \textbf{Planning}}}, \textcolor{ActionColor}{{\fontfamily{lmtt}\selectfont \textbf{Action}}}, and \textcolor{CriticColor}{{\fontfamily{lmtt}\selectfont \textbf{Critic}}}. Please check Section \ref{sec:Daily Mission Clean up} for more discussions. Zoom in for the best view.
	}
\label{fig:example_sr2}
\vspace{-0.1in}
\end{figure*}

\subsubsection{\textit{Honkai: Star Rail}: Daily Mission Clean up Automation}  \label{sec:Daily Mission Clean up}
\noindent\textbf{\textcolor{PlanningColor}{Planning.}} 
The blue captions in Figure \ref{fig:example_sr2} describe the model’s planning phase for automating a daily mission in \textit{Honkai: Star Rail}. The model starts by analyzing the steps needed to accomplish the task: accessing the Interastral Guide, selecting the specific mission (Calyx "Golden for EXP"), adjusting the number of attempts, starting the challenge, activating the auto-battle mode, and finally exiting after the battle concludes. Each step is carefully designed based on the game’s interface and the expected interactions, showcasing the model’s understanding of the game’s task flow, and strictly following the sequence required to automate this routine mission.

\noindent\textbf{\textcolor{ActionColor}{Action.}} In Step 1, the model presses the "Escape" key to open the game menu and navigates to the Interastral Guide. In Step 3, it selects "Calyx Golden for EXP" and then locates and clicks on the "Bud of Memories" entry to begin the mission setup. In Step 5, the model increases the number of challenge attempts to 6 by clicking the "+" button multiple times. It then initiates the challenge by clicking "Start Challenge" in Step 6. Once the battle starts, the model activates auto-battle mode by clicking the icon in Step 7. It waits for the battle to complete, periodically taking screenshots to check the battle’s progress. Finally, after confirming that the mission is complete, the model exits the challenge in Step 11.

\noindent\textbf{\textcolor{CriticColor}{Critic.}} After each action, the model confirms its current state, ensuring the tageting steps executed as planned. For instance, after initiating the auto-battle in Step 7, it monitors the battle’s progress through periodic checks, observing visual cues (like "Battle Start" indicators) to determine the stage of completion. Following the completion screen in Step 11, the model verifies that the task has concluded successfully, indicating that the mission has been fully automated and completed. This feedback process demonstrates the model’s capacity to adapt its actions based on in-game visual feedback, confirming each step’s success in such a long trajectory task.

This example mainly showcases the model’s adeptness in navigating complex game interface, handling a long trajectory task that require multi-step interactions. The model successfully automates a daily mission routine, setting parameters for attempts, and monitoring the battle’s progress, all while ensuring consistency with the user’s request. This capability highlights the model’s great potential in aiding complicated or repetitive gaming tasks that blend strategy, automation, and real-time evaluation.

%% file: sections/5-discussion.tex
% \newpage

\section{Discussion} \label{Discussion}

\subsection{Error Categorization}

We present some representative failure cases in the evaluation, as in Section \ref{sec:Fox Sports}, \ref{sec:Update Name and Phone}, \ref{sec:Insert Numbering}, and \ref{sec:Insert a Sum}. These cases highlight specific areas where the model actions did not align with the user's intended outcomes, revealing limitations in its task comprehension and/or execution. Though the error that causes the failure of the task is versatile, we propose to categorize the presenting error based on our evaluation aspects, explicitly into three sources: \textbf{Planning Error (PE)}, \textbf{Action Error (AE)} and \textbf{Critic Error (CE)}. These categories with examples may help in systematically identifying the natural cause of each failure:

\begin{enumerate}[leftmargin=13pt, itemsep=0pt]
    \item \textbf{Planning Error: } Planning errors occur when the model generates an incorrect plan from task queries, often due to misinterpretation of the task instructions or incorrect current computer state understanding. For example, {\fontfamily{lmtt}\selectfont Task - Fox Sports Subscription}, Figure \ref{fig:example_web3}.
    
    \item \textbf{Action Error: }
    Action errors occur when the agent fails to perform the correct action, given the plan itself is accurate. These errors often relate to the inability in interface understanding, spatial recognition, or precise control within the GUI environment. For example, {\fontfamily{lmtt}\selectfont Task - Insert a Sum Equation over Cells}, Figure \ref{fig:example_excel2}.
    
    \item \textbf{Critic Error: }
    Critic errors occur when the agent incorrectly assesses its own actions or computer state, leading to erroneous action completion feedback. For example,  {\fontfamily{lmtt}\selectfont Task - Update Name and Phone Number on Resume Template}, Figure \ref{fig:example_word3}, and {\fontfamily{lmtt}\selectfont Task - Insert Numbering Symbol}, Figure \ref{fig:example_ppt3}.
    
\end{enumerate}

% TODO: Discussion

% \begin{figure}[h]
% \centering
% \centerline{\includegraphics[width=0.5\linewidth]{figures/error_analysis/error_cate.pdf}}
% \caption[Caption for LOF]{
% TODO: MORE EXPERIMENTS NEED!! Representative categories of failure cases in Computer Use ability evaluation. 
% 	}
% \label{fig:error_cate}
% \vspace{-0.1in}
% \end{figure}

% \subsection{Failure Case Showcase}

\subsection{Toward Future GUI Agents}

\paragraph{Future benchmarking API-based Computer Use models.}
For future benchmarks, there is a critical need for more dynamic and interactive environments that accurately reflect real-world complexities, e.g., different versions of software as the providers update. Moreover, we find that the screen resolution is vital for GUI agents, and this diversity can be considered in the future. Current static datasets and limited interaction paradigms restrict the assessment of an agent's adaptability and capacity to respond to real-world applications.

\paragraph{Critic error correction.} 
Our evaluations reveal that the model frequently misjudges task completion, particularly assuming a task has been completed. This tendency highlights a shortfall in the model’s self-assessment mechanisms. Although some of these problems can be addressed through prompting, a complete solution to this still may require improvements to the GUI agent framework, such as an internalized strict critic module.

\paragraph{Discrepancy with real human computer use.} 
Current model still fail to fully replicate the nuanced human computer use, for example, page scrolling and rigorous browsing. An obvious drawback is that page scrolling based on `Page Up/Down` shortcuts loses a huge portion of the coherence, resulting in fragmented or incomplete interface information. These discrepancies are largely attributed to limitations in training data, which may not fully capture the variability and context-specific adaptations seen in human users.

%% file: sections/6-conclusion.tex
\section{Conclusion}

In this study, we presented a preliminary case study of API-based GUI agent, \textbf{Claude 3.5 Computer Use}, focusing on its performance across diverse desktop environments, including web navigation, workflow, productivity software, and video games. Our case study highlights both the potential and limitations of the current model, particularly in the aspects of planning, action execution, and critic feedback. By providing an out-of-the-box framework, \href{https://github.com/showlab/computer_use_ootb}{Computer Use Out-of-the-Box}, we aim to bridge the accessibility gap to seamlessly deploy and benchmark these models in real-world scenarios. We hope our framework and evaluation approach will contribute to the foundation for further advancements in GUI agent research, driving progress toward more sophisticated and reliable automated computer use models.

%% file: main.bbl
\begin{thebibliography}{10}

\bibitem{liu2023llava}
Haotian Liu, Chunyuan Li, Qingyang Wu, and Yong~Jae Lee.
\newblock Visual instruction tuning.
\newblock In {\em Neural Information Processing Systems}, 2023.

\bibitem{zhu2023minigpt}
Deyao Zhu, Jun Chen, Xiaoqian Shen, Xiang Li, and Mohamed Elhoseiny.
\newblock Minigpt-4: Enhancing vision-language understanding with advanced large language models.
\newblock {\em arXiv preprint arXiv:2304.10592}, 2023.

\bibitem{ye2023mplug}
Qinghao Ye, Haiyang Xu, Guohai Xu, Jiabo Ye, Ming Yan, Yiyang Zhou, Junyang Wang, Anwen Hu, Pengcheng Shi, Yaya Shi, et~al.
\newblock mplug-owl: Modularization empowers large language models with multimodality.
\newblock {\em arXiv preprint arXiv:2304.14178}, 2023.

\bibitem{li2023otter}
Bo~Li, Yuanhan Zhang, Liangyu Chen, Jinghao Wang, Jingkang Yang, and Ziwei Liu.
\newblock Otter: A multi-modal model with in-context instruction tuning.
\newblock {\em arXiv preprint arXiv:2305.03726}, 2023.

\bibitem{wang2023visionllm}
Wenhai Wang, Zhe Chen, Xiaokang Chen, Jiannan Wu, Xizhou Zhu, Gang Zeng, Ping Luo, Tong Lu, Jie Zhou, Yu~Qiao, et~al.
\newblock Visionllm: Large language model is also an open-ended decoder for vision-centric tasks.
\newblock {\em arXiv preprint arXiv:2305.11175}, 2023.

\bibitem{bai2023qwen}
Jinze Bai, Shuai Bai, Shusheng Yang, Shijie Wang, Sinan Tan, Peng Wang, Junyang Lin, Chang Zhou, and Jingren Zhou.
\newblock Qwen-vl: A frontier large vision-language model with versatile abilities.
\newblock {\em arXiv preprint arXiv:2308.12966}, 2023.

\bibitem{chen2023minigpt}
Jun Chen, Deyao Zhu, Xiaoqian Shen, Xiang Li, Zechun Liu, Pengchuan Zhang, Raghuraman Krishnamoorthi, Vikas Chandra, Yunyang Xiong, and Mohamed Elhoseiny.
\newblock Minigpt-v2: large language model as a unified interface for vision-language multi-task learning.
\newblock {\em arXiv preprint arXiv:2310.09478}, 2023.

\bibitem{chen2023shikra}
Keqin Chen, Zhao Zhang, Weili Zeng, Richong Zhang, Feng Zhu, and Rui Zhao.
\newblock Shikra: Unleashing multimodal llm's referential dialogue magic.
\newblock {\em arXiv preprint arXiv:2306.15195}, 2023.

\bibitem{peng2023kosmos}
Zhiliang Peng, Wenhui Wang, Li~Dong, Yaru Hao, Shaohan Huang, Shuming Ma, and Furu Wei.
\newblock Kosmos-2: Grounding multimodal large language models to the world.
\newblock {\em arXiv preprint arXiv:2306.14824}, 2023.

\bibitem{weng2023prompt}
Lilian Weng.
\newblock Llm-powered autonomous agents.
\newblock {\em lilianweng.github.io}, Jun 2023.

\bibitem{sumers2023cognitive}
Theodore~R Sumers, Shunyu Yao, Karthik Narasimhan, and Thomas~L Griffiths.
\newblock Cognitive architectures for language agents.
\newblock {\em arXiv preprint arXiv:2309.02427}, 2023.

\bibitem{renda_survey}
Lei Wang, Chen Ma, Xueyang Feng, Zeyu Zhang, Hao Yang, Jingsen Zhang, Zhiyuan Chen, Jiakai Tang, Xu~Chen, Yankai Lin, Wayne~Xin Zhao, Zhewei Wei, and Ji-Rong Wen.
\newblock A survey on large language model based autonomous agents.
\newblock {\em http://arxiv.org/abs/2308.11432}, 2023.

\bibitem{sun2023push}
Qiushi Sun, Zhangyue Yin, Xiang Li, Zhiyong Wu, Xipeng Qiu, and Lingpeng Kong.
\newblock Corex: Pushing the boundaries of complex reasoning through multi-model collaboration.
\newblock {\em arXiv preprint arXiv:2310.00280}, 2023.

\bibitem{hong2024metagpt}
Sirui Hong, Mingchen Zhuge, Jonathan Chen, Xiawu Zheng, Yuheng Cheng, Jinlin Wang, Ceyao Zhang, Zili Wang, Steven Ka~Shing Yau, Zijuan Lin, Liyang Zhou, Chenyu Ran, Lingfeng Xiao, Chenglin Wu, and J{\"u}rgen Schmidhuber.
\newblock Meta{GPT}: Meta programming for a multi-agent collaborative framework.
\newblock In {\em The Twelfth International Conference on Learning Representations}, 2024.

\bibitem{durante2024agentaisurvey}
Zane Durante, Qiuyuan Huang, Naoki Wake, Ran Gong, Jae~Sung Park, Bidipta Sarkar, Rohan Taori, Yusuke Noda, Demetri Terzopoulos, Yejin Choi, Katsushi Ikeuchi, Hoi Vo, Li~Fei-Fei, and Jianfeng Gao.
\newblock Agent ai: Surveying the horizons of multimodal interaction, 2024.

\bibitem{sun2024survey}
Qiushi Sun, Zhirui Chen, Fangzhi Xu, Kanzhi Cheng, Chang Ma, Zhangyue Yin, Jianing Wang, Chengcheng Han, Renyu Zhu, Shuai Yuan, et~al.
\newblock A survey of neural code intelligence: Paradigms, advances and beyond.
\newblock {\em arXiv preprint arXiv:2403.14734}, 2024.

\bibitem{wu2024copilot}
Zhiyong Wu, Chengcheng Han, Zichen Ding, Zhenmin Weng, Zhoumianze Liu, Shunyu Yao, Tao Yu, and Lingpeng Kong.
\newblock Os-copilot: Towards generalist computer agents with self-improvement.
\newblock {\em arXiv preprint arXiv:2402.07456}, 2024.

\bibitem{zhang2024ufo}
Chaoyun Zhang, Liqun Li, Shilin He, Xu~Zhang, Bo~Qiao, Si~Qin, Minghua Ma, Yu~Kang, Qingwei Lin, Saravan Rajmohan, et~al.
\newblock Ufo: A ui-focused agent for windows os interaction.
\newblock {\em arXiv preprint arXiv:2402.07939}, 2024.

\bibitem{cheng2024seeclick}
Kanzhi Cheng, Qiushi Sun, Yougang Chu, Fangzhi Xu, Yantao Li, Jianbing Zhang, and Zhiyong Wu.
\newblock Seeclick: Harnessing gui grounding for advanced visual gui agents.
\newblock {\em arXiv preprint arXiv:2401.10935}, 2024.

\bibitem{hong2024cogagent}
Wenyi Hong, Weihan Wang, Qingsong Lv, Jiazheng Xu, Wenmeng Yu, Junhui Ji, Yan Wang, Zihan Wang, Yuxiao Dong, Ming Ding, et~al.
\newblock Cogagent: A visual language model for gui agents.
\newblock In {\em Proceedings of the IEEE/CVF Conference on Computer Vision and Pattern Recognition}, pages 14281--14290, 2024.

\bibitem{zheng2024seeact}
Boyuan Zheng, Boyu Gou, Jihyung Kil, Huan Sun, and Yu~Su.
\newblock Gpt-4v(ision) is a generalist web agent, if grounded.
\newblock In {\em Forty-first International Conference on Machine Learning}, 2024.

\bibitem{nakano2021webgpt}
Reiichiro Nakano, Jacob Hilton, Suchir Balaji, Jeff Wu, Long Ouyang, Christina Kim, Christopher Hesse, Shantanu Jain, Vineet Kosaraju, William Saunders, et~al.
\newblock Webgpt: Browser-assisted question-answering with human feedback.
\newblock {\em arXiv preprint arXiv:2112.09332}, 2021.

\bibitem{yin2024agent}
Da~Yin, Faeze Brahman, Abhilasha Ravichander, Khyathi Chandu, Kai-Wei Chang, Yejin Choi, and Bill~Yuchen Lin.
\newblock Agent lumos: Unified and modular training for open-source language agents.
\newblock In {\em Proceedings of the 62nd Annual Meeting of the Association for Computational Linguistics (Volume 1: Long Papers)}, pages 12380--12403, 2024.

\bibitem{lai2024autowebglm}
Hanyu Lai, Xiao Liu, Iat~Long Iong, Shuntian Yao, Yuxuan Chen, Pengbo Shen, Hao Yu, Hanchen Zhang, Xiaohan Zhang, Yuxiao Dong, et~al.
\newblock Autowebglm: Bootstrap and reinforce a large language model-based web navigating agent.
\newblock {\em arXiv preprint arXiv:2404.03648}, 2024.

\bibitem{zhang2023you}
Zhuosheng Zhang and Aston Zhang.
\newblock You only look at screens: Multimodal chain-of-action agents.
\newblock {\em arXiv preprint arXiv:2309.11436}, 2023.

\bibitem{zhang2023appagent}
Chi Zhang, Zhao Yang, Jiaxuan Liu, Yucheng Han, Xin Chen, Zebiao Huang, Bin Fu, and Gang Yu.
\newblock Appagent: Multimodal agents as smartphone users.
\newblock {\em arXiv preprint arXiv:2312.13771}, 2023.

\bibitem{niu2024screenagent}
Runliang Niu, Jindong Li, Shiqi Wang, Yali Fu, Xiyu Hu, Xueyuan Leng, He~Kong, Yi~Chang, and Qi~Wang.
\newblock Screenagent: A vision language model-driven computer control agent.
\newblock {\em arXiv preprint arXiv:2402.07945}, 2024.

\bibitem{gao2023assistgui}
Difei Gao, Lei Ji, Zechen Bai, Mingyu Ouyang, Peiran Li, Dongxing Mao, Qinchen Wu, Weichen Zhang, Peiyi Wang, Xiangwu Guo, et~al.
\newblock Assistgui: Task-oriented desktop graphical user interface automation.
\newblock {\em arXiv preprint arXiv:2312.13108}, 2023.

\bibitem{zhang2024xlam}
Jianguo Zhang, Tian Lan, Ming Zhu, Zuxin Liu, Thai Hoang, Shirley Kokane, Weiran Yao, Juntao Tan, Akshara Prabhakar, Haolin Chen, et~al.
\newblock xlam: A family of large action models to empower ai agent systems.
\newblock {\em arXiv preprint arXiv:2409.03215}, 2024.

\bibitem{zhang2024agentohana}
Jianguo Zhang, Tian Lan, Rithesh Murthy, Zhiwei Liu, Weiran Yao, Juntao Tan, Thai Hoang, Liangwei Yang, Yihao Feng, Zuxin Liu, et~al.
\newblock Agentohana: Design unified data and training pipeline for effective agent learning.
\newblock {\em arXiv preprint arXiv:2402.15506}, 2024.

\bibitem{zeng2023agenttuning}
Aohan Zeng, Mingdao Liu, Rui Lu, Bowen Wang, Xiao Liu, Yuxiao Dong, and Jie Tang.
\newblock Agenttuning: Enabling generalized agent abilities for llms.
\newblock {\em arXiv preprint arXiv:2310.12823}, 2023.

\bibitem{yin2023lumos}
Da~Yin, Faeze Brahman, Abhilasha Ravichander, Khyathi Chandu, Kai-Wei Chang, Yejin Choi, and Bill~Yuchen Lin.
\newblock Lumos: Learning agents with unified data, modular design, and open-source llms.
\newblock {\em arXiv preprint arXiv:2311.05657}, 2023.

\bibitem{li2024ac}
Wei Li, William Bishop, Alice Li, Chris Rawles, Folawiyo Campbell-Ajala, Divya Tyamagundlu, and Oriana Riva.
\newblock On the effects of data scale on computer control agents.
\newblock {\em arXiv preprint arXiv:2406.03679}, 2024.

\bibitem{xu2024envisions}
Fangzhi Xu, Qiushi Sun, Kanzhi Cheng, Jun Liu, Yu~Qiao, and Zhiyong Wu.
\newblock Interactive evolution: A neural-symbolic self-training framework for large language models.
\newblock {\em arXiv preprint arXiv:2406.11736}, 2024.

\bibitem{gou2024uground}
Boyu Gou, Ruohan Wang, Boyuan Zheng, Yanan Xie, Cheng Chang, Yiheng Shu, Huan Sun, and Yu~Su.
\newblock Navigating the digital world as humans do: Universal visual grounding for gui agents.
\newblock {\em arXiv preprint arXiv:2410.05243}, 2024.

\bibitem{koh2024tree}
Jing~Yu Koh, Stephen McAleer, Daniel Fried, and Ruslan Salakhutdinov.
\newblock Tree search for language model agents.
\newblock {\em arXiv preprint arXiv:2407.01476}, 2024.

\bibitem{yao2023react}
Shunyu Yao, Jeffrey Zhao, Dian Yu, Nan Du, Izhak Shafran, Karthik Narasimhan, and Yuan Cao.
\newblock React: synergizing reasoning and acting in language models (2022).
\newblock {\em arXiv preprint arXiv:2210.03629}, 2023.

\bibitem{anthropic2024claude35}
{Anthropic}.
\newblock Introducing computer use, a new claude 3.5 sonnet, and claude 3.5 haiku, October 2024.
\newblock Accessed: 2024-11-02.

\end{thebibliography}
